%% file: arXiv.tex
\documentclass[10pt,twocolumn,letterpaper]{article}

\usepackage[algorithms]{wacv}      


\definecolor{wacvblue}{rgb}{0.21,0.49,0.74}
\usepackage[pagebackref,breaklinks,colorlinks,allcolors=wacvblue]{hyperref}

\usepackage{algorithm}
\usepackage{algorithmic}
\usepackage{multirow}
\usepackage[table]{xcolor}
\usepackage{xcolor}

\def\confName{WACV}
\def\confYear{2027}

\newcommand{\parhead}[1]{%
  \vspace{0.25em}\noindent\textbf{#1}\;
}

\title{MINER: Modeling Inter- and Intra-Person Variance for Reidentification}

\author{Oron Nir\\
Reichman University \& Microsoft\\
{\tt\small oron.nir@post.runi.ac.il}
\and
Eliyahu Strugo\\
Microsoft\\
{\tt\small elistrugo@microsoft.com}
\and
Ariel Shamir\\
Reichman University\\
{\tt\small arik@runi.ac.il}
}

\begin{document}
\maketitle
\input{sec/0_abstract}

\input{sec/1_intro}
\input{sec/2_related_work}

\input{sec/3_method}
\input{sec/4_balancing_intr}
\input{sec/5_experiments}

\input{sec/6_discussion}
\input{sec/8_conclussion}

\input{sec/10_acknowledgements}
{
    \small
    \bibliographystyle{ieeenat_fullname}
    \bibliography{main}
}
\clearpage
\appendix

\twocolumn[{
    \centering
    {\LARGE\bfseries
    Supplemental Material for\par}
    \vspace{0.3em}
    {\Large\bfseries
    \textit{MINER: Modeling Inter- and Intra-Person Variance for Reidentification}\par}
    \vspace{1.5em}
}]

\input{sec/supplement_sections}

\end{document}

%% file: sec/0_abstract.tex
\begin{abstract}
Clothes-Changing Person Re-Identification (CC-ReID) remains challenging due to drastic appearance shifts across outfits.
Most prior works devise biometric signal-extraction heuristics to enhance inter-person variance, e.g., using segmentation masks, while leaving intra-person variance uncontrolled.
We present \textbf{MINER}, a simple model-agnostic framework that suppresses garment bias by contrasting hard examples and adaptively balancing dataset-dependent intra- and inter-person variance.
Extensive experiments show that MINER improves retrieval across CC-ReID benchmarks, achieving state-of-the-art performance on four datasets and ranking among the leaders in all three tracks of the open TVRID RGB-Depth ReID challenge.
For code and checkpoints see \href{https://oronnir.github.io/MINER/}{oronnir.github.io/MINER}.
\end{abstract}

%% file: sec/1_intro.tex
\section{Introduction}
\label{sec:introduction}

Person re-identification (ReID) aims to retrieve images of the same individual across disjoint camera views and has become a central component in modern large-scale surveillance, robotics, and video analytics systems.
In this setting, a query is matched against a gallery of images from other cameras, and performance is evaluated by identity ranking.
While standard ReID benchmarks assume fixed clothing, real-world deployments often involve drastic outfit changes over time. 
This gives rise to the more challenging task of \emph{Clothes-Changing Person Re-Identification (CC-ReID)}, where appearance variations induced by clothing largely dominate the visual signal.
Consequently, CC-ReID requires models to rely on subtle biometric cues, e.g., body shape features, while remaining invariant to garments.

Most prior CC-ReID methods explicitly aim to suppress clothing-related features and at the same time enhance biometric cues by either disentangling biometric and appearance factors, leveraging auxiliary clothing parsers, or attribute supervision. 
These approaches focus primarily on modeling \emph{inter-person variance}—namely, maximizing the discrepancy between different identities.
While this design philosophy is highly effective for improving clothing invariance, it does not take into account \emph{intra-person variance} directly. Namely, it minimizes the discrepancy within different outfits of the same identity as a byproduct of metric learning rather than as a first-class modeling objective.

In this work, we revisit this fundamental assumption and demonstrate that \emph{simultaneously controlling the balance of within-person and between-person variance} leads to substantial gains in CC-ReID performance.
We show that explicitly concentrating the training process on visually similar yet distinct identities, while simultaneously enforcing maximal cross-clothing variability within each identity, produces significantly more discriminative and robust embedding spaces.
This dual optimization principle naturally yields strong improvements in global retrieval quality and state-of-the-art (SOTA) performance on multiple CC-ReID benchmarks, including PRCC, LTCC, Celeb-ReID-light (Celeb-L), and DeepChange.

\textbf{MINER} is an inter-class identity-level and intra-class clothing-level stochastic hard mining strategy that can be used to finetune existing models, enhancing their performance. 
MINER constructs training batches by (i) soft-sampling visually similar but different identities as hard negatives using a reference encoder, and (ii) soft-sampling maximally dissimilar clothing pairs cross-camera as hard positives. 
Such controlled yet stochastic exposure to extreme intra- and inter-identity appearance variations drives a contrastive optimization process that explicitly aligns with the goal of CC-ReID.
We further formulate a multi-positive contrastive objective, which leverages block-diagonal identity supervision across query–gallery pairs during training.
In summary, our contributions are:
\vspace{-0.1mm}
\begin{itemize}
    \item We formulate CC-ReID as a trade-off between inter-person discrimination and cross-clothing invariance, and derive an online controller that adapts it from gradient variance---without grid search or extra forward passes.
    \item We combine stochastic hard mining with multi-positive contrastive learning over hard negative identities and cross-clothing positives.
    \item MINER improves SOTA backbones on four CC-ReID benchmarks and ranks top-3 in all TVRID 2026 tracks.
\end{itemize}

%% file: sec/2_related_work.tex
\section{Related Work}
\label{sec:related}

\parhead{Person Re-Identification.}
Person ReID aims to match images of the same individual across non-overlapping cameras, often under large pose, illumination, and viewpoint changes.
Early deep ReID methods leverage convolutional backbones with identification loss functions and triplet-style metric learning, building strong baselines that remain competitive and widely used~\cite{zheng2015scalable,zheng2017unlabeled,sun2018beyond,wang2018learning,luo2019bag}.
Recent works have shifted towards vision transformer-based architectures (ViT), most notably TransReID~\cite{he2021transreid}.
Follow-up surveys highlight the importance of the global receptive field and multi-granularity features for improving context through its inductive bias~\cite{ye2024transformer_for_object_reid_survey,gu2025reidmamba}.
Large-scale pre-training and self-supervised objectives further improve generalization: \citet{chen2023beyond} perform semantic controllable self-supervised pre-training for ReID using Swin backbones, while \citet{fu2022lupersonnl} leverage noisy tracker-based labels at scale to learn robust features using ResNet-50.
Our method is backbone-agnostic and therefore applicable to CNN-, ViT-, Swin-based, and any other architecture.

\parhead{Clothes-Changing Re-Identification.}
CC-ReID focuses on long-term identity tracking, where garments vary dramatically between observations of the same person. Dedicated datasets, like PRCC~\cite{yang2020prcc}, expose extreme appearance gaps, and have motivated a rich line of methods to isolate clothes-agnostic cues~\cite{wan2020changing,yu2020cocas,gu2022cal,wang2023exploring}.
A large body of work exploits structural priors: body-shape~\cite{hong2021fgsam}, silhouette~\cite{jin2022gireid}, pose~\cite{liu2024pose_guided}, human parsing~\cite{shu2021semantic,ding2025fa}, head or face fusion~\cite{shi2022iranet}, and co-attention mechanisms to disentangle body structure from clothing appearance~\cite{yang2023cloth_debiasing}.

Recent approaches rely on visual semantic modeling and CLIP-driven representations~\cite{radford2021learning}.
CLIP-ReID~\cite{li2023clip} and others adapt contrastive vision-language pretraining~\cite{zhou2022coop} to ReID, while CCAF, MMPL, and MADE use masked or prompt-based text to highlight identity-relevant semantics and suppress clothing information~\cite{li2024ccaf,osamura2024mmpl,peng2024made}.
Cloth-Changing Aware Transformer (CCAT) injects cloth-aware priors into a transformer backbone to improve identity robustness across outfits~\cite{ren2022ccat}.
In parallel, identity/disentanglement approaches such as DIFFER~\cite{liang2025differ}, GEFF~\cite{arkushin2024geff}, and \citet{chen2024frd} factor biometric identity from non-biometric attributes (clothes, hairstyle, accessories) via gradient reversal or face/body fusion, achieving strong CC-ReID performance.
However, CC-ReID remains an open problem, as clothing changes substantially hinder identity discrimination. Most existing methods treat query–gallery pairs independently during training, leaving direct contrastive modeling of visually similar identities across clothing variations largely underexplored.


\parhead{Cross-Modal ReID.}
Security applications drive increasing interest in different night-time ReID signals~\cite{jiang2022crossmodal,josi2023multimodal,qiu2024hos,qiu2025hoh}.
RGB-Depth maps RGB query to Depth gallery~\citep{delecluse2026icpr}.
While these tasks use a same-clothes setup, we use it to demonstrate the generality of our method to factor out the cross-modality gap instead.

\parhead{Hard-Sample Mining in (CC-)ReID.}
Hard-sample mining prioritizes the most difficult positive and negative examples under the current embedding to strengthen discriminative learning.
Classic hard triplet losses and point-to-set variants, e.g., \citet{hermans2017defense}, emphasize difficult positives and negatives within each batch to sharpen the embedding geometry \cite{yu2018hard,smirnov2018hard,zhou2023adaptive}.
\citet{sheng2020globalhard} propose global hard sample mining, arguing that mining from the whole training set yields harder samples than local batch mining and can further improve performance.
\citet{yan2021beyond} suggest a fine-grained difference-aware pairwise loss to better capture subtle appearance differences in traditional ReID.

In CC-ReID, hard mining has been explored more recently.
HSGL (Hard Sample Generation and Learning) introduces a multimodal framework that \emph{generates} hard samples guided by both textual and visual semantics to emphasize clothing-induced appearance variance~\cite{liu2025try_harder_hsgl}.
Other works implicitly mine hard examples through cloth-aware augmentation, pose-diversified synthesis, or cross-domain clothes-changing scenarios \cite{zhang2023MBUNet,guo2023SCNet,yang2023acid,liu2024cloth,kim2024posedive}.

Our MINER sampler performs \emph{centroid-level} hard mining within precomputed person and clothing clusters: it jointly selects hard-negative identities (highly similar centroids across person identities) and hard-positive clothing pairs (with a low similarity across outfits for the same person identity) using a soft sampling distribution.
This yields batches with controlled intra-/inter-person difficulty while remaining compatible with standard contrastive training.

\parhead{Contrastive Learning for CC-ReID.}
Contrastive learning has become a central paradigm for representation learning in both supervised and self-supervised settings~\cite{oord2018cpc,chen2020simclr,he2020moco}.
In vision-language models such as CLIP, symmetric contrastive objectives over image-text pairs yield highly transferable representations, motivating their adoption in CC-ReID via dual-encoder training~\cite{li2024ccaf}.
Recent CC-ReID methods extend this paradigm by adapting CLIP-style supervision to emphasize biometric cues while mitigating clothing-induced bias~\cite{osamura2024mmpl,peng2024made,liang2025differ}.
However, despite relying on a single-positive contrastive setup, CLIP fundamentally depends on large batches to provide sufficient implicit negatives, without which optimization becomes unstable and prone to representation collapse~\cite{chen2020simclr}.

Instead, we reformulate contrastive learning over multi-positive hard-mined sampled batches, where positives and negatives are organized across identities, cameras, and clothing sets.
By applying an InfoNCE-style loss over these structured sets, MINER enables targeted soft hard-mining that focuses gradient updates on challenging intra-person outfit changes and inter-person look-alikes.

%% file: sec/3_method.tex
\section{Method}
\label{sec:method}

{\looseness=-1
To refine a model for clothes-agnostic identity representation we define a sampling mechanism that controls identity and clothing selection per batch.
The main idea is to contrast visually similar distinct identities where each identity is depicted in visually different appearances. We sample identities over their embeddings centroids and gather them into a bipartite batch denoted as a query set and a gallery set.
\par}
\subsection{MINER Sampler}
\label{subsec:miner_sampler}
Let $\mathcal{I}$ be a set of training images, each image $x \in \mathcal{I}$ is annotated by a tuple $(p,c,a)$, where
$p \in \mathcal{P}$ is the person identity (PID),
$c \in \mathcal{C}$ is the camera index,
and $a \in \mathcal{A}$ is the clothing ID.
Let $f(\cdot)$ denote a pre-trained image encoder to be finetuned.
We pre-compute the following centroids:
\begin{equation}
\label{eq:centroids_update}
    \mathbf{z}_p
    = \frac{1}{|\mathcal{I}_p|}
      \sum_{x \in \mathcal{I}_p} f(x),
    ~~~
    \mathbf{z}_{p,c,a}
    = \frac{1}{|\mathcal{I}_{p,c,a}|}
      \sum_{x \in \mathcal{I}_{p,c,a}} f(x),
\end{equation}
where $\mathcal{I}_p$ is the set of all images of person $p$,
and $\mathcal{I}_{p,c,a}$ the subset of images of $(p,c,a)$.

\parhead{Camera-pair candidate identity set.}
Given a pair of distinct cameras $c_q \neq c_g$, we define the set of
valid candidate person identities, $\mathcal{P}_{c_q,c_g}$, which appear in both cameras with at least two different clothing configurations, i.e.,
\begin{equation}
\label{eq:camera_pairs}
\begin{aligned}
\mathcal{P}_{c_q,c_g}
=\{& \, p \in \mathcal{P} \;\mid\;\exists\, a_q \neq a_g: \\
&\ \mathcal{I}_{p,c_q,a_q} \neq \emptyset,
\ \mathcal{I}_{p,c_g,a_g} \neq \emptyset \,\}
\end{aligned}
\end{equation}

\parhead{Soft Hard-Negative PID Mining.}
We compose a batch of $m$ person identities by first uniformly sampling one
\begin{equation*}
    p_s \sim \mathcal{U}\bigl(\mathcal{P}_{c_q,c_g}\bigr),
\end{equation*}
and computing cosine similarity in $f$-space
$\forall p\! \in\! \mathcal{P}_{c_q,c_g}$:
\vspace{-1pt}
\begin{equation}
\label{eq:cosine_sim_pids}
    s(p, p_s)
    =
    \frac{\mathbf{z}_p^\top \mathbf{z}_{p_s}}
         {\|\mathbf{z}_p\|_2 \, \|\mathbf{z}_{p_s}\|_2}.
\end{equation}
For all $p \in \mathcal{P}_{c_q,c_g}$, we define
\vspace{-1pt}
\begin{equation}
\label{eq:pid_sample_distribution}
\pi(p \mid p_s)
=
\frac{\exp\big(s(p,p_s)\big)}
{\sum_{p' \in \mathcal{P}_{c_q,c_g}} \exp\big(s(p',p_s)\big)}.
\end{equation}
Rather than taking $m$ nearest identities deterministically---brittle under noisy centroids---the batch identity set $\mathcal{B}$ is obtained by sampling $m$ identities proportional to $\pi(p\! \mid\! p_s)$ without replacement, thereby increasing inter-person variance by emphasizing (soft) hard-negative identities.

\parhead{Soft Hard-Positive Clothing-Pair Mining.}
After selecting visually similar identities, we construct query–gallery pairs that increase intra-person variation across clothing and camera views.
For each $p\! \in\! \mathcal{B}$, the valid clothing pairs are:
\begin{equation}
\label{eq:soft_pos_mining_clothes}
\mathcal{A}_{p}^{c_q,c_g}\!=\!\{(a_q, a_g) \!\mid\! a_q \!\in\! \mathcal{A}_{p,c_q},
a_g \!\in\! \mathcal{A}_{p,c_g}, 
\ a_q \!\neq\! a_g \}
\end{equation}
where $\mathcal{A}_{p,c}$ is the set of clothing IDs worn by $p$ in camera $c$.
Each pair $(a_q, a_g)$ is scored via cosine similarity between their
$f$ centroids
\begin{equation}
    s(a_q, a_g \mid p)
    =
    \frac{
        {\mathbf{z}_{p,c_q,a_q}}^\top \mathbf{z}_{p,c_g,a_g}
    }{
        \|\mathbf{z}_{p,c_q,a_q}\|_2
        \,\|\mathbf{z}_{p,c_g,a_g}\|_2
    }.
\end{equation}
Rather than selecting the hardest clothing pair deterministically, we sample
the cross-camera clothing pair stochastically using a soft argmin distribution.
For each $p \in \mathcal{B}$ and each $(a_q, a_g) \in \mathcal{A}_{p}^{c_q,c_g}$,
we define
\begin{equation}
\pi(a_q, a_g \mid p)
=
\frac{
\exp(- s(a_q, a_g \mid p))
}{
\sum_{(a'_q,a'_g)}
\exp(- s(a'_q, a'_g \mid p))
}
\end{equation}
where $(a'_q,a'_g)\in \mathcal{A}_{p}^{c_q,c_g}$. The clothing pair $(a_q^*, a_g^*)$ is then sampled as
\begin{equation}
\label{eq:sampled_clothes_pair}
(a_q^*, a_g^*) \sim \pi(a_q, a_g \mid p).
\end{equation}

\parhead{Final training tuples.}
For each $p \in \mathcal{B}$, we sample
$n$ images from $\mathcal{I}_{p,c_q,a_q^*}$,
$n$ images from $\mathcal{I}_{p,c_g,a_g^*}$,
and use them as query and gallery images with shared identity $p$.

\begin{algorithm}[t]
\caption{MINER Hard Mining for CC-ReID}
\begin{algorithmic}[1]
\REQUIRE Set $\mathcal{I}$; Encoder $f$;
         Batch parameters $\langle m, n\rangle$
\ENSURE Query set $\mathcal{Q}$, gallery set $\mathcal{G}$
\STATE Sample distinct cameras $(c_q, c_g)$.
\STATE Construct the candidate identity set
       $\mathcal{P}_{c_q,c_g}$ (Eq.~\ref{eq:camera_pairs}).
\STATE Sample one identity uniformly $p_s \sim \mathcal{U}(\mathcal{P}_{c_q,c_g})$.
\STATE $\forall p \in \mathcal{P}_{c_q,c_g}$ compute similarity $s(p, p_s)$ (Eq.~\ref{eq:cosine_sim_pids}).
\STATE Draw $m$ identities without replacement by $\pi(p \mid p_s)$.
\STATE Initialize $\mathcal{Q} \leftarrow \emptyset$, $\mathcal{G} \leftarrow \emptyset$.
\FORALL{$p \in \mathcal{B}$}
    \STATE Construct $\mathcal{A}_{p}^{c_q,c_g}$ for different clothes (Eq.~\ref{eq:soft_pos_mining_clothes}).
    \STATE Sample hard clothing pair $(a_q^*, a_g^*)$ (Eq.~\ref{eq:sampled_clothes_pair}).
    \STATE Sample $n$ images from $\mathcal{I}_{p,c_q,a_q^*}$
           and add them to $\mathcal{Q}$.
    \STATE Sample $n$ images from $\mathcal{I}_{p,c_g,a_g^*}$
           and add them to $\mathcal{G}$.
\ENDFOR
\STATE Update centroids $\mathbf{z}_p,~\mathbf{z}_{p,c,a}$ (Eq.~\ref{eq:centroids_update}).
\STATE \textbf{return} $\mathcal{Q}, \mathcal{G}$.
\end{algorithmic}
\label{alg:miner_sampller}
\end{algorithm}

\subsection{Contrastive Loss Function}
\label{subsec:contrastive_loss}

\begin{figure}[t]
\centering
\includegraphics[width=1.025\linewidth,trim=4.85cm 1.35cm 4.85cm 0cm, 
    clip]{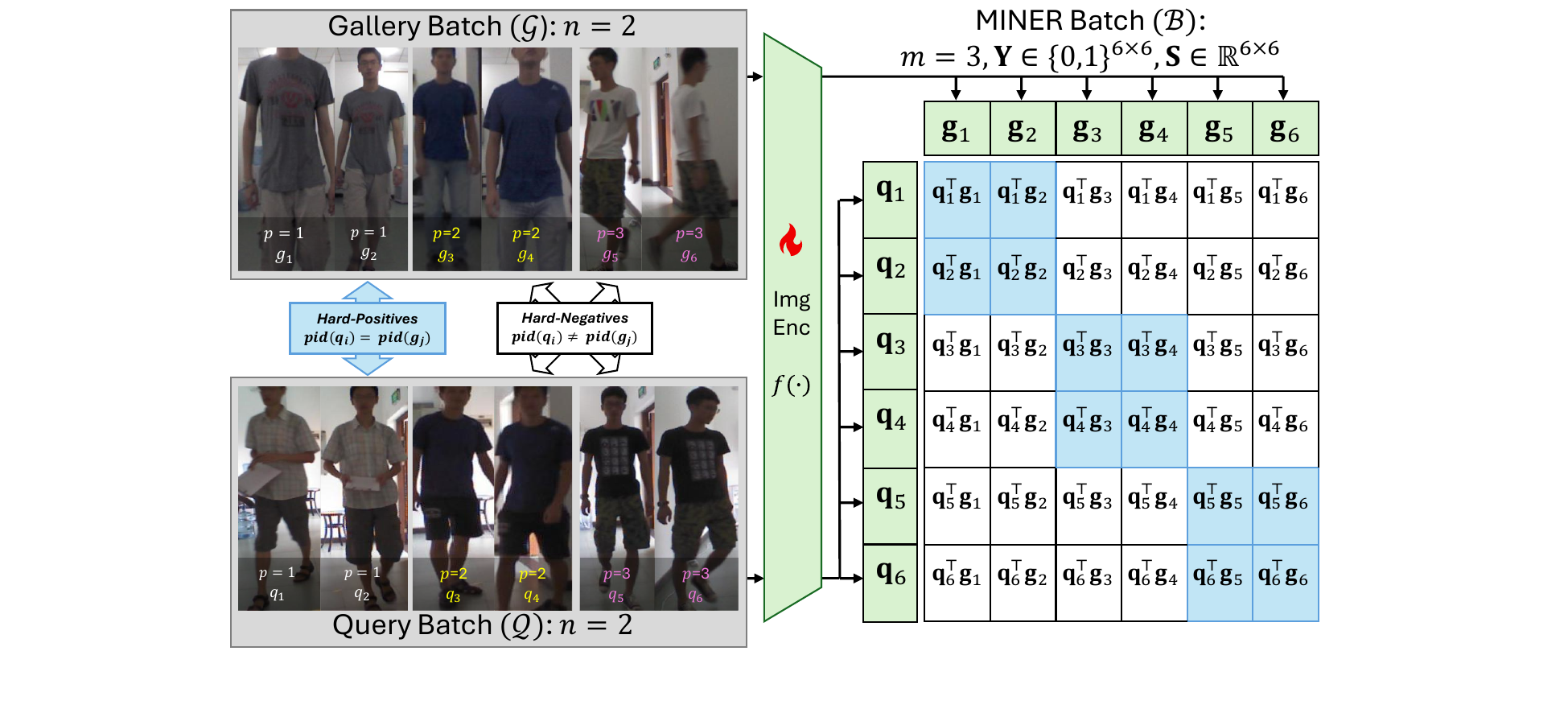}
\caption{MINER Contrastive Loss over a bipartite sampled batch. Hard-Negative person IDs and Hard-Positive examples per person distill both inter-person (white) and intra-person (blue) variance simultaneously using a block-diagonal label matrix.}
\label{fig:teaser_contrastive_learning}
\label{fig:miner_hard_mining}
\end{figure}

Given a mined batch produced by Alg.~\ref{fig:miner_hard_mining}, we obtain
a set of query images $\mathcal{Q}=\{q_i\}_{i=1}^{N}$ and a set of gallery images
$\mathcal{G}=\{g_j\}_{j=1}^{N}$, where $N = m n$, so each identity contributes
$n$ query images and $n$ gallery images.
Recall that $f(\cdot)$ denotes the CC-ReID encoder producing $\ell_2$-normalized embeddings,
we define the query and gallery embeddings as $\mathbf{q}_i = f(q_i)~\text{and} ~ \mathbf{g}_j = f(g_j)$, respectively.

\parhead{Similarity matrix.}
We compute the dense pairwise similarity matrix
$\mathbf{S}\! \in\! \mathbb{R}^{N \times N}$ as: $S_{ij} = \mathbf{q}_i^\top \mathbf{g}_j,$
 like CLIP~\cite{radford2021learning}.

\parhead{Block-diagonal ground-truth structure.}
By construction, each identity $p \in \mathcal{B}$ contributes exactly $n$ query samples and $n$ gallery samples per batch. 
Therefore, the binary ground-truth correspondence matrix $\mathbf{Y} \in \{0,1\}^{N \times N}$ assumes a block-diagonal structure like SupCon~\cite{khosla2020supervised}:
\vspace{-1pt}
\begin{equation}
Y_{ij} = 
\begin{cases}
1, & \text{if } \mathrm{pid}(q_i) = \mathrm{pid}(g_j), \\
0, & \text{otherwise},
\end{cases}
\end{equation}
where each diagonal block of shape $n\times n$ represents all cross-clothing
matches between queries and galleries of the same identity within the batch.
All off-diagonal blocks correspond to hard negative cross-identity matches (see illustration of $m=3,n=2$ in Fig.~\ref{fig:teaser_contrastive_learning}).  

\begin{figure*}[htbp]
\begin{minipage}[t]{0.49\linewidth}
        \centering
        \includegraphics[width=\textwidth,trim=0.25cm 0.25cm 0.25cm 0.25cm, 
    clip]{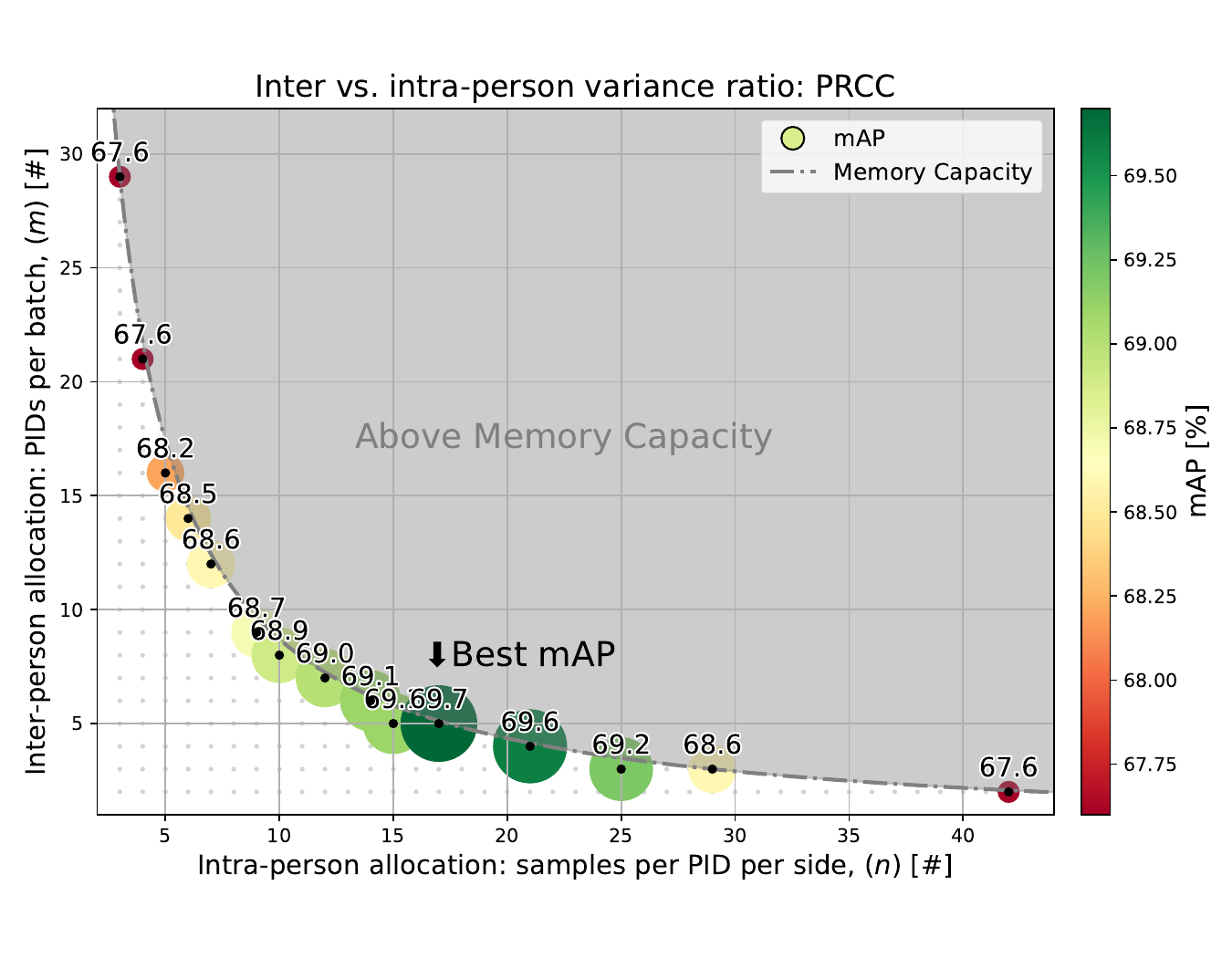}
        \caption{
        The mAP as a function of intra- and inter-person examples per batch, within memory constraints.
        The performance peaks at $m=5$ PIDs and $n=17$ images per person on PRCC.
        }
        \label{fig:balanceMemCapacityPRCC}
\end{minipage}
\hfill
\begin{minipage}[t]{0.49\linewidth}
    \centering
        \centering
        \includegraphics[width=\textwidth,trim=0.25cm 0.25cm 0.25cm 0.25cm, 
    clip]{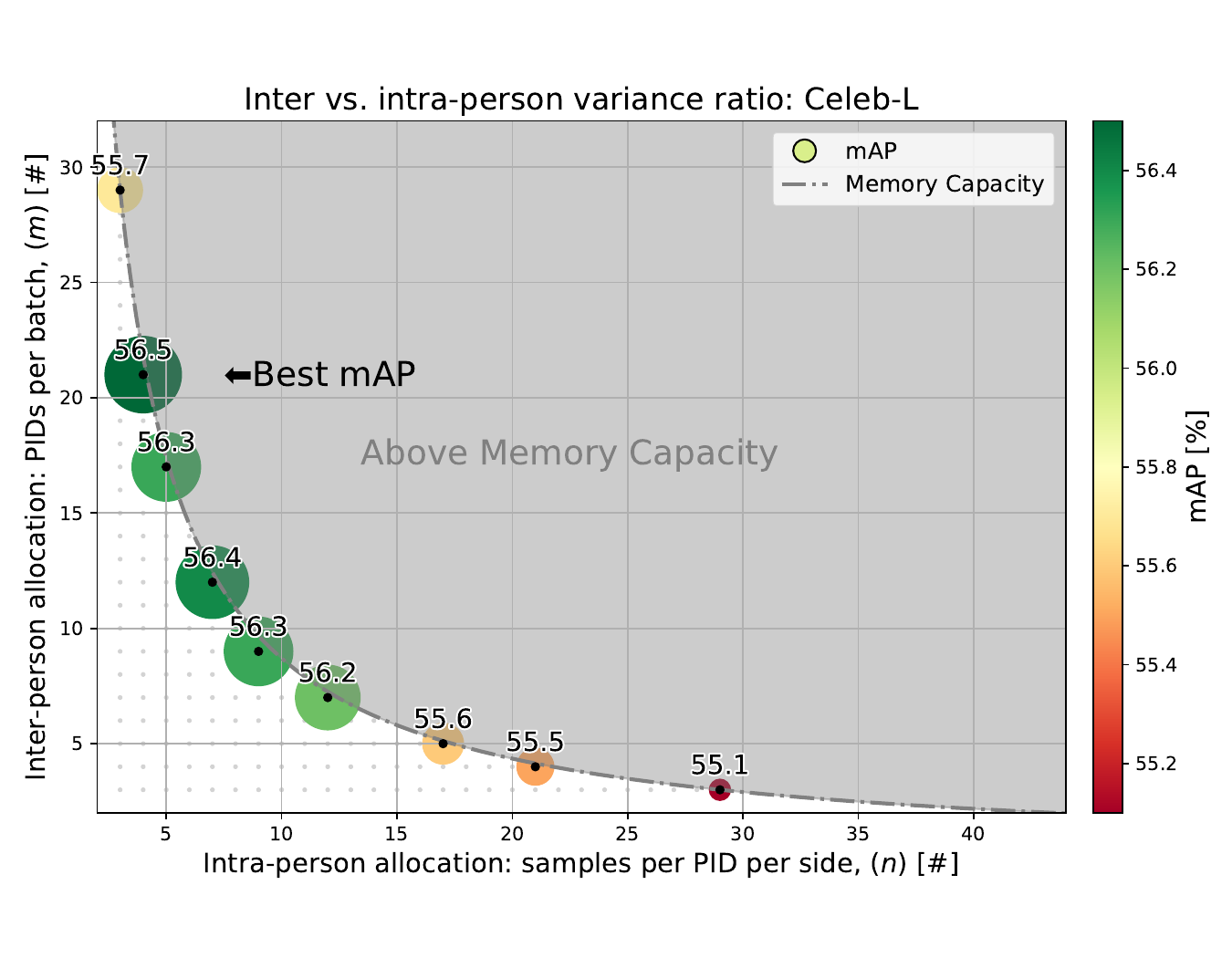}
        \caption{The empirically best batch composition differs substantially across datasets: mAP peaks for Celeb-L at $m=21$ PIDs and $n=4$ images per person, motivating dataset-adaptive allocation.}
        \label{fig:balanceMemCapacity2CelebL}
\end{minipage}
\end{figure*}

\parhead{Contrastive Objective.}
We adopt a symmetric cross-entropy contrastive objective over rows and columns
of $\mathbf{S}$, adapting contrastive loss to the CC-ReID setting. The query-to-gallery
loss is defined as
\vspace{-1pt}
\begin{equation}
\label{eq:loss_q2g}
\mathcal{L}_{q \rightarrow g}
=
-\frac{1}{N n}
\sum_{i=1}^{N}
\sum_{j=1}^{N}
Y_{ij}
\log
\frac{
\exp(S_{ij})
}{
\sum_{k=1}^{N} \exp(S_{ik})
},
\end{equation}
and the gallery-to-query loss is defined symmetrically as
\vspace{-1pt}
\begin{equation}
\label{eq:loss_g2q} 
\mathcal{L}_{g \rightarrow q}
=
-\frac{1}{N n}
\sum_{j=1}^{N}
\sum_{i=1}^{N}
Y_{ij}
\log
\frac{
\exp(S_{ij})
}{
\sum_{k=1}^{N} \exp(S_{kj})
}.
\end{equation}
The final training objective is $\mathcal{L}_{\text{total}}
=
\frac{1}{2}
\left(
\mathcal{L}_{q \rightarrow g}
+
\mathcal{L}_{g \rightarrow q}
\right)$.
Unlike simple contrastive training, where only a single positive pair exists per row,
our block-diagonal supervision induces multiple positive targets per query and per
gallery sample. This formulation naturally enforces:
(i) cross-clothing invariance within identity blocks, and
(ii) strong discrimination against visually similar hard negatives mined at the
identity level. As a result, the learned embedding space is explicitly optimized for clothing-agnostic cross-camera identity retrieval.

\subsection{Adaptive Batch Composition}
\label{sec:adaptive_var}
{\looseness=-1
Subsections~\ref{subsec:miner_sampler}--\ref{subsec:contrastive_loss} built a sampler and loss that,
for a given batch shape $\langle m,n\rangle$, expose controlled inter- and
intra-person variance; that shape has so far been hand-set, and we now set it
automatically. Under a fixed image budget $B=2mn\le B_{\max}$, $m$ and $n$
compete---more identities means fewer images per identity---so the design choice
is not the \emph{size} of the batch, but the \emph{allocation} of $B_{\max}$
between inter-person diversity and intra-person cross-clothing coverage.
As we show in \cref{sec:balance_analysis}, this is a genuine, dataset-dependent trade-off. 
The two knobs serve different objectives: $m$ stabilizes discrimination, which is shared across the batch; $n$ stabilizes cross-clothing alignment, which must be estimated per identity. 
The objective currently estimated less reliably deserves more budget. 
We thus cast batch composition as a budget-constrained estimation problem, and minimize the resulting
risk under $2mn\le B_{\max}$. Its minimizer is closed-form and depends on the data
through a single scalar---the ratio of inter- to intra-person gradient variance---which the
backward pass on $\mathcal{L}_{\text{total}}$ already yields, so the allocation
adapts online with no extra forward pass and no grid search.
\par}

\parhead{Two structurally distinct gradient signals.}
The contrastive objective separates exactly into a within-block (positive) and a
between/off-block (negative) part. Differentiating the per-query loss $\ell_i$
through the softmax, propagating through $\nabla_\theta S_{ik}$, and summing the
symmetric gallery-to-query term (standard softmax cross-entropy) decomposes the batch
gradient of $\mathcal{L}_{\text{total}}$ into two additive signals,
\vspace{-1pt}
\begin{equation}
\label{eq:grad_split}
h \;=\; \underbrace{h_{\mathrm{inv}}}_{\text{within-block (positives)}}
       \;+\;
       \underbrace{h_{\mathrm{disc}}}_{\text{off-block (negatives)}} ,
\end{equation}
the \emph{invariance} signal $h_{\mathrm{inv}}$ that collapses cross-clothing
positives of the same identity, and the \emph{discrimination} signal
$h_{\mathrm{disc}}$ that separates visually similar hard negatives. By
construction $h_{\mathrm{inv}}$ exists only when each identity supplies at least
one cross-clothing pair ($n\ge 1$, $a_q^*\neq a_g^*$), and is refined by
\emph{more images per identity} $(n)$; $h_{\mathrm{disc}}$ exists only when
$m\ge 2$, and is refined by \emph{more identities} $(m).$ The two knobs thus act
on two different gradient directions, and as we now show, the budget refines them
at different rates.

\parhead{Variance components of a two-stage mined batch.}
MINER draws a batch in two stages: $m$ identities from $\pi(p\mid p_s)$
(Eq.~\ref{eq:pid_sample_distribution}), then $2n$ images per identity under the
mined clothing pair (Eq.~\ref{eq:sampled_clothes_pair})---a classical two-stage
sampling design~\cite{cochran1977sampling}, whose gradient estimator admits a
between/within variance-components decomposition~\cite{searle1992variance}. With
per-identity mean gradient $\phi_p=\mathbb{E}_{x\sim p}[\nabla_\theta\ell(x)]$ and
its mean $\mu=\mathbb{E}_p[\phi_p]$, define the \emph{between-identity} and
\emph{within-identity} covariances
\begin{equation}
\label{eq:var_components}
\Sigma_{\mathrm{B}}=\operatorname{Cov}_p\!\big[\phi_p\big],
\quad
\Sigma_{\mathrm{W}}=\mathbb{E}_p\!\big[\operatorname{Cov}_{x\sim p}[\nabla_\theta \ell(x)]\big].
\end{equation}
$\operatorname{tr}\Sigma_{\mathrm{B}}$ is the dispersion of the descent direction
\emph{across} identities; $\operatorname{tr}\Sigma_{\mathrm{W}}$, its dispersion
\emph{within} an identity. The two enter the batch at different sample sizes,
$m$ and $2n$.

\parhead{Budget-constrained allocation.}
Under a fixed budget, each signal is governed by the variance of its own
estimator. Discrimination concerns the \emph{shared} descent direction $\mu$, of
which every batch identity is an observation; pooling the $m$ per-identity means
estimates $\mu$ with trace-variance
$\frac{\operatorname{tr}\Sigma_{\mathrm{B}}}{m} + \frac{\operatorname{tr}\Sigma_{\mathrm{W}}}{2mn}$.
The second term equals $\frac{\operatorname{tr}\Sigma_{\mathrm{W}}}{B_{\max}}$, a
constant under the budget, so only $\frac{\operatorname{tr}\Sigma_{\mathrm{B}}}{m}$
is allocation-relevant. Invariance is \emph{not} shared: by the block-diagonal
structure of $\mathbf{Y}$, positive pairs lie strictly within identity blocks, so
collapsing identity $p$'s outfits requires $\phi_p$ estimated from $p$'s own $2n$
images alone---images of other identities never enter $p$'s positive pairs and
cannot sharpen it. Its rate is the \emph{average per-identity} estimation error
$\mathbb{E}_p\|\hat{\phi}_p-\phi_p\|_2^2=\frac{\operatorname{tr}\Sigma_{\mathrm{W}}}{2n}$,
in which $m$ does not appear: we average $m$ local errors of size
$\frac{\operatorname{tr}\Sigma_{\mathrm{W}}}{2n}$ rather than estimating a common
mean---the shared-representation versus task-specific rate split of multi-task
learning~\cite{baxter2000model,maurer2016benefit}, each identity acting as a
task. The allocation-relevant risk is their sum,
\vspace{-1pt}
\begin{equation}
\label{eq:risk}
\mathcal{E}(m,n)=\frac{\operatorname{tr}\Sigma_{\mathrm{B}}}{m}
+
\frac{\operatorname{tr}\Sigma_{\mathrm{W}}}{2n},
\quad \text{s.t.}\quad 2mn = B_{\max}.
\end{equation}
The first term is the allocation-relevant variance of the pooled
(population-descent) gradient; minimizing it alone collapses the allocation to
$n=1$. This collapse is the blind spot of a single-pooled-gradient
view~\cite{mccandlish2018empirical}: it is insensitive to whether \emph{each}
identity is well aligned, whereas cross-clothing retrieval is per-identity by
construction---mAP averages per-query AP, each query fixed by its own
identity's geometry, a quantity other identities cannot sharpen. Eq.~\ref{eq:risk}
therefore adds the second term---the per-identity alignment the pooled view
omits---making the inter/intra allocation a genuine trade-off rather than a
degenerate ``maximize $m$.'' We do not claim $\mathcal{E}$ tracks mAP pointwise;
rather, its closed-form optimum recovers the grid-search mAP optimum of
Figs.~\ref{fig:balanceMemCapacityPRCC}--\ref{fig:balanceMemCapacity2CelebL},
under which MINER attains the results of \cref{sec:experiments}. Minimizing
$\mathcal{E}$ along the budget frontier ($\mathcal{E}$ is strictly convex in $n$)
gives the unique allocation
\vspace{-1pt}
\begin{equation}
\label{eq:waterfill}
\boxed{
\begin{aligned}
n^{\star}
&= \sqrt{\,B_{\max}\,\frac{\operatorname{tr}\Sigma_{\mathrm{W}}}
{4\operatorname{tr}\Sigma_{\mathrm{B}}}}\,,
\;
m^{\star}
= \sqrt{\,B_{\max}\,\frac{\operatorname{tr}\Sigma_{\mathrm{B}}}
{\operatorname{tr}\Sigma_{\mathrm{W}}}}\,,
\\[0.5em]
 \rho&:=\frac{m^{\star}}{n^{\star}}=\frac{2\operatorname{tr}\Sigma_{\mathrm{B}}}
{\operatorname{tr}\Sigma_{\mathrm{W}}}\,.
\end{aligned}
}
\end{equation}
The constraint holds identically, $2m^{\star}n^{\star}=B_{\max}$, and the optimum
depends on the data through the \emph{single scalar} $\rho$, the ratio of inter-
to intra-person gradient variance. The symmetry of $\mathcal{L}_{\text{total}}$
in query-gallery fixes an equal per-side split; $\langle m^\star,n^\star\rangle$ is projected
to feasible integers $\langle m,n\rangle$.

\parhead{Modeling assumptions.}
Eq.~\ref{eq:waterfill} rests on three modeling choices and affects only the adaptive selection of $\langle m,n\rangle$, not the contrastive objective itself.
\emph{(i)} We take the allocation objective to be the equally weighted sum of the two estimation risks above; their $\frac{1}{m}$ and $\frac{1}{2n}$ scalings come from the corresponding estimator variances, so the lone modeling choice is the equal weighting, which rescales $\rho$ but not its functional form.
\emph{(ii)} Covariances are evaluated under the conditional mining (Alg.~\ref{alg:miner_sampller}), not IID: the between-ID term under $\pi(p\mid p_s)$ and the within-ID term under $\pi(a_q,a_g\mid p)$; we assume the between-ID dispersion is carried mainly by $h_{\mathrm{disc}}$ and the within-ID dispersion mainly by $h_{\mathrm{inv}}$, i.e. their cross-covariance is small relative to the marginal traces, so the two risks decouple as in Eq.~\ref{eq:risk}.
\emph{(iii)} Projection-head gradients provide a stable low-variance proxy for the corresponding full-parameter covariance traces. We address these assumptions in the supplement.

\parhead{Online variance-component estimators.}
Both traces in Eq.~\ref{eq:waterfill} are estimable from the current mined batch
via one-way random-effects ANOVA. With per-identity means
$\hat{\phi}_p = \tfrac{1}{2n}\sum_{x\in p}\nabla_\theta\ell(x)$ and grand mean
$\hat{\mu}=\tfrac{1}{m}\sum_p \hat{\phi}_p$, the ANOVA-style variance-component estimators are
\vspace{-1pt}
\begin{align}
\label{eq:est_W}
\widehat{\operatorname{tr}\Sigma}_{\mathrm{W}}
&= \frac{1}{m(2n-1)}\sum_{p}\sum_{x\in p}\big\|\nabla_\theta\ell(x)-\hat{\phi}_p\big\|_2^2 ,\\
\label{eq:est_B}
\widehat{\operatorname{tr}\Sigma}_{\mathrm{B}}
&= \frac{1}{m-1}\sum_{p}\big\|\hat{\phi}_p-\hat{\mu}\big\|_2^2
   \;-\; \frac{1}{2n}\,\widehat{\operatorname{tr}\Sigma}_{\mathrm{W}},
\end{align}
the second term in Eq.~\ref{eq:est_B} removing the within-identity noise that
inflates the raw between-identity scatter. Both reduce to sums of \emph{squared
gradient norms} grouped by identity, accumulated on-device in a single
squared-norm read per identity, mirroring \citet{mccandlish2018empirical} but
resolved per identity: the $m$ in Eqs.~\ref{eq:est_W}--\ref{eq:est_B} sharpens
the population-trace estimate, not the per-identity alignment error. To keep the cost negligible we evaluate the norms on the projection
head only---the gradient w.r.t.\ the $\ell_2$-normalized embedding
$\nabla_{\mathbf{z}}\ell$, available with no additional backward pass and a
standard low-variance surrogate for the full-parameter gradient. The two traces
are smoothed by an exponential moving average (EMA) across steps,
$\bar{\Sigma}_{\bullet}\!\leftarrow\!(1-\eta)\bar{\Sigma}_{\bullet}+\eta\,\widehat{\operatorname{tr}\Sigma}_{\bullet}$,
to suppress single-batch noise. The overall variance controller is detailed in
Alg.~\ref{alg:var_controller}.

\begin{algorithm}[t]
\caption{Variance Controller}
\begin{algorithmic}[1]
\REQUIRE Memory budget $B_{\max}$; EMA rate $\eta$; interval $T$
\STATE Initialize $\langle m, n\rangle$ from a feasible default; $\bar{\Sigma}_{\mathrm{B}},\bar{\Sigma}_{\mathrm{W}}\!\leftarrow\!0$
\FOR{each training step $t$}
    \STATE Mine batch with current $\langle m,n\rangle$ (Alg.~\ref{alg:miner_sampller}).
    \STATE Backprop $\mathcal{L}_{\text{total}}$.
    \STATE Form $\widehat{\operatorname{tr}\Sigma}_{\mathrm{W}},\widehat{\operatorname{tr}\Sigma}_{\mathrm{B}}$ (Eqs.~\ref{eq:est_W}--\ref{eq:est_B}).
    \STATE $\bar{\Sigma}_{\bullet}\!\leftarrow\!(1-\eta)\bar{\Sigma}_{\bullet}+\eta\,\widehat{\operatorname{tr}\Sigma}_{\bullet}$
    \IF{$t \bmod T = 0$}
        \STATE $n^\star\!\leftarrow\!\sqrt{\frac{B_{\max}\,\bar{\Sigma}_{\mathrm{W}}}{4\bar{\Sigma}_{\mathrm{B}}}}$;\ \ $m^\star\leftarrow \frac{B_{\max}}{2n^\star}$
        \STATE Project $\langle m^\star,n^\star\rangle$ to feasible integers $\langle m,n\rangle$.
    \ENDIF
\ENDFOR
\end{algorithmic}
\label{alg:var_controller}
\end{algorithm}

%% file: sec/4_balancing_intr.tex
\section{Balancing the Batch Variance}
\label{sec:balance_analysis}
Both intra- and inter-person variance are central to CC-ReID representation, but
\emph{are they equally important?} We probe this on PRCC by sweeping the allocation
$\langle m,n\rangle$ along the memory-capacity
frontier ($B=2mn\le174$) and recording mAP (Fig.~\ref{fig:balanceMemCapacityPRCC}).
mAP rises to a single peak at $\langle 5,17\rangle$ ($69.7\%$) and declines
on either side: the $m$-heavy allocation $\langle 29,3\rangle$ and the
$n$-heavy $\langle 2,42\rangle$ fall to $67.6\%$.
PRCC is captured under controlled conditions, so its identities are mutually
similar and $\operatorname{tr}\Sigma_B$ is small; the resulting low $\rho$ ($0.31$) favors larger $n$.

\begin{figure}[t]
\centering
\includegraphics[width=1\linewidth]{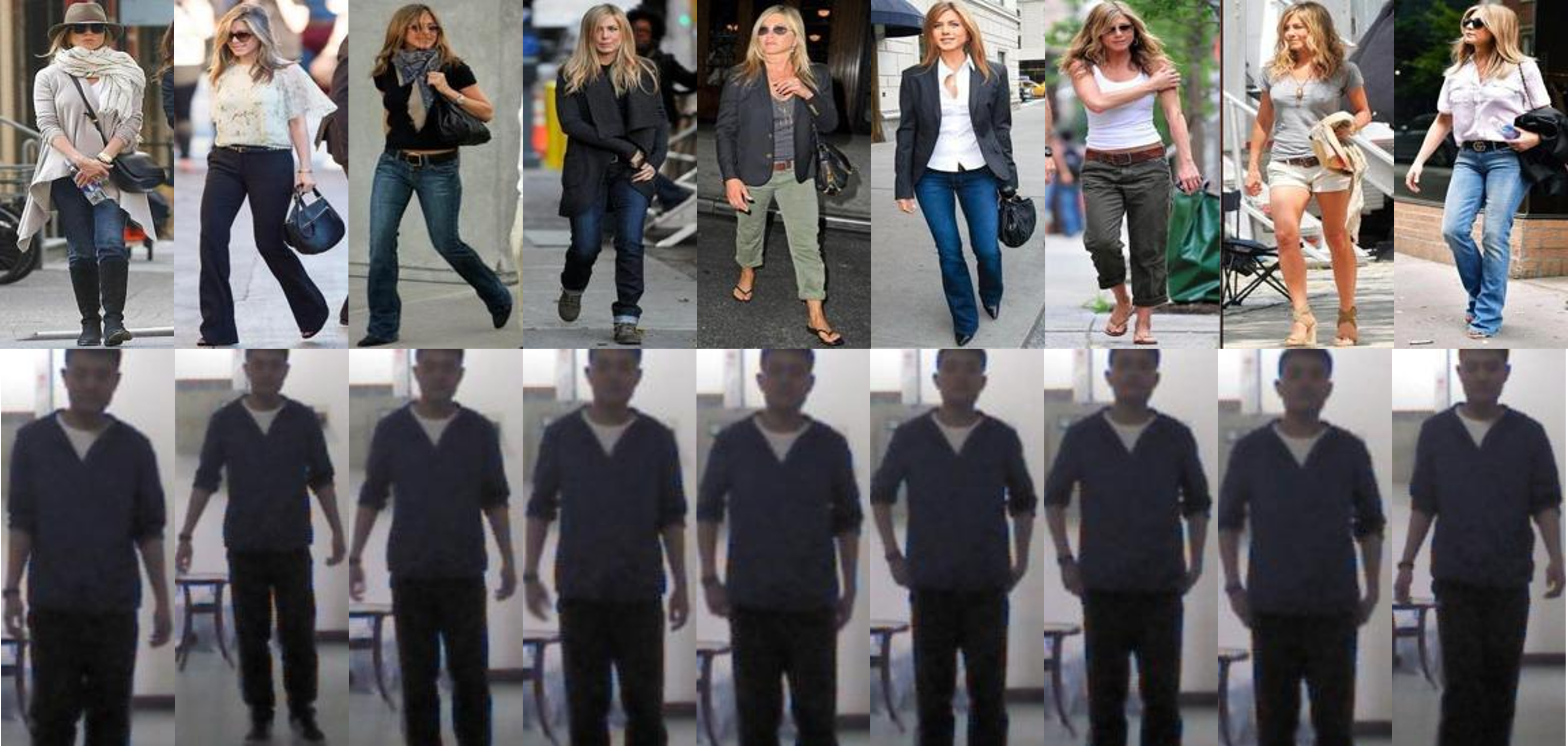}
\caption{The same identity exhibits greater intra-person variance in Celeb-L (top) than
in PRCC (bottom), spanning clothing, pose, and scene.}
\label{fig:prcc_and_celebs}
\end{figure}

To show how this balance shifts with the data prior, we repeat the analysis on Celeb-L,
collected ``in-the-wild'' through paparazzi photos (Fig.~\ref{fig:prcc_and_celebs}).
Both components rise---no identity recurs in the same outfit, and the identities themselves are far more heterogeneous---but $\operatorname{tr}\Sigma_B$ rises faster: $\rho$ grows to $5.34$, so Celeb-L peaks at a higher $\frac{m}{n}$ ratio: $\langle 21,4\rangle$, i.e., more identities and fewer samples each (Fig.~\ref{fig:balanceMemCapacity2CelebL}).

\parhead{Does the controller earn its complexity?}
Adaptive allocation is warranted if (i)~no single $\langle m,n\rangle$ is optimal
across datasets and (ii)~the controller recovers each dataset's optimum without grid
search; both hold. The PRCC and Celeb-L optima lie at opposite ends of the frontier and
do not transfer: $\langle 5,17\rangle$ loses $0.9$ mAP on Celeb-L, $\langle 21,4\rangle$
loses $2.1$ on PRCC. Since this prior is unknown at training time, the relevant criterion
is worst-case minimax regret~\cite{savage1951}: the controller incurs only $0.2$ mAP
worst-case and $0.10$ mean---lower on both than every fixed split (best fixed: $0.6$ and
$0.33$, on different splits)---thus Pareto-dominating all nine (see supplement). 
It recovers each optimum at negligible cost: regret $0.0/0.1/0.2$ on PRCC/LTCC/Celeb-L.

%% file: sec/5_experiments.tex
\section{Experiments}
\label{sec:experiments}
We evaluate MINER, with its Variance Controller, and report Rank@1 and mAP metrics on six CC-ReID datasets, and two traditional ReID protocols: PRCC~\cite{yang2020prcc}, LTCC~\cite{qian2020ltcc}, Celeb-L~\cite{huang2021celeb}, DeepChange~\cite{xu2023deepchange}, VC-Clothes~\cite{wan2020changing}, LaST~\cite{shu2021last}. 
MINER was also submitted to three different tracks of the open competition TVRID~\cite{delecluse2026icpr}. 
We follow the evaluation protocol of full dataset-isolation, i.e., training a new model per dataset and splitting training from test with no re-ranking.
For evaluation protocols and implementation details, see the supplemental material.

\parhead{CC-ReID Finetuning Evaluation.}
We present our main results in Tab.~\ref{tab:miner_results} over the datasets above where MINER is compared with 17 methods \citep{gu2022cal,shu2021last,zhang2023MBUNet,yang2023acid,guo2023SCNet,li2024ccaf,peng2024made,liu2024clip3dreid,yin2025rflnet,ding2025fa,pathak2025colors,ding2025ptreid,hao2025masked,liang2025differ,jonnalagedda2025novel,yu2026hpl,kim2026ortho}.
We initiate the finetune with pretrained checkpoints and run 20 epochs with Alg.~\ref{alg:var_controller}.
The experiments demonstrate MINER using EVA-CLIP~\cite{sun2023eva} and DIFFER~\cite{liang2025differ} checkpoints separately.
Our method finetunes backbone models and achieves SOTA scores on 4 datasets; VC/LaST remain competitive.
The controller estimated $\langle m^\star,n^\star\rangle=\langle5.14,16.54\rangle$ on PRCC and $\langle21.30,3.99\rangle$ on Celeb-L, closely matching the validation grid optima $\langle5,17\rangle$ and $\langle21,4\rangle$, respectively, from \cref{sec:balance_analysis} ($B_{\max}=170$).
The same validation was conducted on LTCC, resulting in three near-tied mAP peaks. The controller converged to one of them: the continuous estimate $\langle 9.38,9.06\rangle$, and a grid-based near-peak $\langle 9,9\rangle$.
Thus, the Variance Controller recovers the dataset-specific allocation without grid search.
\begin{table*}[t]
\begin{center}
\begin{small}
\begin{sc}
\centering
\resizebox{\linewidth}{!}{
\begin{tabular}{lllcccccccccccc}
\toprule
\textbf{\small{CC-ReID}} 
& \textbf{\small{Backbone}}
& \textbf{\small{Vanilla}}
& \multicolumn{2}{c}{\textbf{\small{PRCC}}} 
& \multicolumn{2}{c}{\textbf{\small{LTCC}}}  
& \multicolumn{2}{c}{\textbf{\small{Celeb-L}}}
& \multicolumn{2}{c}{\textbf{\small{DeepChange}}} 
& \multicolumn{2}{c}{\textbf{\small{VC-Clothes}}} 
& \multicolumn{2}{c}{\textbf{\small{LaST}}} \\
\textbf{\small{Method}}& & \small{\textbf{Weights}} & \small{R@1} & \small{mAP} & \small{R@1} & \small{mAP} & \small{R@1} & \small{mAP} & \small{R@1} & \small{mAP} & \small{R@1} & \small{mAP} & \small{R@1} & \small{mAP}\\
\midrule
\small{CAL~\citep{gu2022cal}} \tiny{`22} & ResNet50 & $\checkmark$     & 55.2 & 55.8 & 40.1 & 18.0 & -- & -- & {54.0} & {19.0} & 81.4 & 81.7 & 73.7 & 37.2  \\
\small{LaST~\citep{shu2021last}} \tiny{`22} & ResNet50 & $\checkmark$     & 57.7 & 54.7 & -- & -- & 54.4 & 11.8 & -- & -- & -- & --  & 71.0 & 28.0 \\
\small{MBUNet~\citep{zhang2023MBUNet}} \tiny{`23} & ResNet50 &   & 68.7 & 65.2 & 40.3 & 15.0 & 35.5 & 21.5 & -- & -- & 82.7 & 70.3 & -- & --  \\
\small{ACID~\citep{yang2023acid}} \tiny{`23} &   ResNet50 &$\checkmark$  & 55.4 & {66.1} & 29.1 & 14.5 & 27.9 & 15.8 & -- & -- & 84.3 & 74.2 & -- & --  \\
\small{SCNet~\citep{guo2023SCNet}} \tiny{`23}&  ResNet50 & $\checkmark$  & 58.7 & 59.9 & 47.5 & 25.5 & -- & -- & 53.5 & 18.7 & -- & -- & -- & --  \\
\small{CCAF~\citep{li2024ccaf}}  \tiny{`23} & ViT-B/CLIP & $\checkmark$  & {70.4} & 63.7 & 42.9 & 20.1 & -- & -- & -- & -- & -- & -- & -- & --  \\
\small{MADE~\citep{peng2024made}}  \tiny{`24} & ViT-L/EVA & $\checkmark$  & 64.3 & 59.1 & 47.4 & 24.4 & 72.0 & 52.3 & -- & -- & -- & --  & {79.0} & {40.9} \\
\small{CLIP3DReID~\citep{liu2024clip3dreid}}  \tiny{`24} & ResNet50 & $\checkmark$  & 60.6 & 59.3 & 42.1 & 21.7 & 39.4 & 26.3 & 56.7 & 20.8 & -- & -- & -- & -- \\
\small{RFLNet~\citep{yin2025rflnet}} \tiny{`25}& ResNet50 & $\checkmark$   & 58.1 & 59.0 & 43.9 & 22.2 & -- & -- & 52.4 & {19.0} & \textbf{89.1} & 82.9 & -- & --  \\
\small{FA-ReID~\citep{ding2025fa}} \tiny{`25} &   ResNet50 & $\checkmark$   & 63.7 & 60.7 & 43.9 & 19.9 & -- & -- & -- & -- & 85.7 & \underline{85.9} & -- & --  \\
\small{CSCI~\citep{pathak2025colors}} \tiny{`25} &ViT-L/EVA & $\checkmark$& 66.2 & 61.3 & 47.8 & 24.4 & -- & -- & -- & -- & -- & --  & -- & -- \\
\small{PT-ReID~\citep{ding2025ptreid}} \tiny{`25}& ViT-B/CLIP & $\checkmark$ & 66.3 & 63.4 & 48.0 & 25.0 & -- & -- & -- & -- & 86.8 & \textbf{86.5}  & -- & -- \\
\small{MTAT~\citep{hao2025masked}} \tiny{`25}& ViT-L/EVA & $\checkmark$ & -- & -- & 48.0 & 22.3 & 71.0 & 51.4 & -- & -- & -- & --  & -- & -- \\
\small{DIFFER~\citep{liang2025differ}} \tiny{`25} & ViT-L/EVA & $\checkmark$  & 68.5 & 64.7 & {58.2} & {31.6} & {75.6} & {54.3} & -- & -- & -- & -- & -- & --  \\
\small{HPL~\citep{yu2026hpl}} \tiny{`25}& ResNet50 & $\checkmark$ & {74.3} & 69.0 & -- & -- & -- & -- & -- & -- & 84.9 & 84.7  & -- & -- \\
\small{ORB~\citep{jonnalagedda2025novel}} \tiny{`26} & ResNet50 & & 62.0 & 58.3 & 43.1 & 22.1 & -- & -- & -- & -- & {88.2} & 85.8  & -- & -- \\
\small{Ortho~\citep{kim2026ortho}} \tiny{`26} & ViT-L/EVA & $\checkmark$ & \underline{74.4} & \textbf{70.2} & 56.1 & 30.2 & \textbf{79.1} & \underline{59.0} & -- & -- & -- & --  & \textbf{84.3} & \textbf{53.5} \\
\midrule
\small{MINER \tiny{(DIFFER)}}& ViT-L/EVA&& 73.9 & {69.7} & \textbf{60.9} & \underline{33.3} & {76.9} & {56.3} & \underline{57.8} & \underline{20.0} & \underline{88.8} & 79.8 & {82.3} & {49.5}  \\
\rowcolor{blue!10}
\small{MINER \tiny{(EVA-CLIP)}}& ViT-L/EVA&$\checkmark$ & \textbf{74.5} & \underline{70.0} & \underline{60.2} & \textbf{34.4} & \underline{77.2} & \textbf{59.1} & \textbf{66.8} & \textbf{27.7} & 81.3 & 82.0 & \underline{83.0} & \underline{50.4}  \\

\bottomrule
\end{tabular}
}
\end{sc}
\end{small}
\end{center}
\vspace{-0.8em}
\caption{Comparison of various methods on CC-ReID: MINER Rank@1 and mAP performance across CC-ReID datasets vs. SOTA. We initialize to both EVA-CLIP and DIFFER separately. All methods use pre-trained weights, Vanilla Weights were trained on generic datasets like ImageNet while others used additional ReID finetuning. VC-Clothes is synthetic. Best result is in \textbf{bold}, the second is \underline{underlined}.}
\label{tab:miner_results}
\end{table*}

\parhead{MINER CC-ReID vs. Backbones.}
To verify the impact of our method on different backbones, like CLIP-ReID~\cite{li2023clip} and MADE~\cite{peng2024made}, for the CC-ReID task we reproduce and evaluate our method on PRCC, LTCC, and Celeb-L. 
CLIP-ReID base underperforms as it was not trained for CC-ReID.
MINER improves all backbones in all datasets by $6.7$ and $4.6$ Rank@1 and mAP points, respectively (\cref{tab:miner_ablation}).
Furthermore, MINER gains $(33.3, 32.5)$, $(6.5, 5.6)$ $(\Delta \overline{R@1}, \Delta \overline{mAP})$ over vanilla pretrained EVA-CLIP and ID-Loss baseline respectively on PRCC, LTCC, and Celeb-L. See full analysis in the supplement.
\begin{table}[t]
\begin{center}
\begin{small}
\begin{sc}
\centering
\resizebox{\linewidth}{!}{
\begin{tabular}{llcccc}
\toprule
\textbf{\small{CC-ReID}}     &\small{Data} & \multicolumn{2}{c}{\textbf{\small{Base-reproduced}}}  & \multicolumn{2}{c}{\textbf{\small{+MINER}}}  \\
\textbf{\small{Method}}      & \small{set} & \small{R@1} & \small{mAP} & \small{R@1} & \small{mAP}\\
\midrule
\small{CLIP-ReID}  & \tiny{PRCC} & 17.5 & 18.0 & \textbf{29.0} & \textbf{30.3} \\
\small{MADE}      & \tiny{PRCC} & 61.7  & 56.2 & \textbf{71.2} & \textbf{64.7}\\
\small{DIFFER}    & \tiny{PRCC} & 68.5  & 64.7 & \textbf{73.9} & \textbf{69.7}\\
\midrule
\small{CLIP-ReID}  & \tiny{LTCC} & 17.3 & 8.3  & \textbf{32.7} & \textbf{11.2}\\
\small{MADE}      & \tiny{LTCC} & 44.0 & 19.4 & \textbf{47.4} & \textbf{23.3}\\
\small{DIFFER}    & \tiny{LTCC} & 58.2 & 31.6 & \textbf{60.9} & \textbf{33.3}\\
\midrule
\small{CLIP-ReID}  & \tiny{Celeb-L} & 12.0 & 6.4 & \textbf{21.0} & \textbf{10.3} \\
\small{MADE}      & \tiny{Celeb-L} & 66.0  & 48.4 & \textbf{68.0} & \textbf{49.5}\\
\small{DIFFER}    & \tiny{Celeb-L} & 75.6 & 54.3 & \textbf{76.9} & \textbf{56.3}\\
\bottomrule
\small{\textbf{Average}}     &              & 46.8 & 34.1 & \textbf{53.4} & \textbf{38.7} \\
\small{\textbf{Gain (\(\Delta\))}}    &              &      &      & \textbf{+6.7}  & \textbf{+4.6} \\
\end{tabular}
}
\end{sc}
\end{small}
\end{center}
\vspace{-0.8em}
\caption{Gains from using MINER on various reproduced backbone scores in Rank@1 and mAP across three CC-ReID datasets. MINER improves both metrics on all backbones and datasets.}
\label{tab:miner_ablation}
\end{table}

{\looseness=-1
\parhead{Standard ReID.}
While our work focuses on CC-ReID, traditional ReID is still a critical task in CCTV applications.
We conducted a Same-Clothes (SC) traditional evaluation on PRCC and LTCC (General) against seven methods~\citep{he2021transreid,gu2022cal,peng2024made,yin2025rflnet,ding2025fa,pathak2025colors,liang2025differ}.\
Although MINER suppresses clothing-related cues, it remains effective in SC-ReID, achieving SOTA on both datasets with (Rank@1, mAP) of $(100.0,99.4)$ and $(93.8,67.7)$, respectively (supplement).
\par}

\parhead{Cross-Modal ReID.}
We extend MINER to cross-modal learning.
We adapt MINER by addressing the modality indicator per-source (e.g., RGB or depth) instead of clothes-ID indicator $\mathcal{I}_{p,c,a}$. 
By contrasting hard-negatives across identities and hard-positive across modalities, the algorithm controls inter- and intra-person variance like in CC-ReID.
MINER was submitted to the ICPR 2026 Competition on Privacy-Preserving Person Re-Identification from Top-View RGB-Depth Camera (TVRID)~\cite{delecluse2026icpr}. The challenge addressed three top-view ReID challenges: RGB-RGB, Depth-Depth, and RGB-Depth. 
The setup is challenging as faces are not seen visually through the overhead sensor installation. 
With 148 competing teams and 732 submissions, MINER was the sole method to reach the podium in all three tracks, with (Rank@1, mAP) of $(99.8,99.9), (97.2,98.3), (91.5,94.9)$, respectively.


%% file: sec/6_discussion.tex
\section{Discussion}
\label{sec:discussion}
Most CC-ReID work treats intra-person variance across outfits as a byproduct of metric learning, prioritizing inter-person separation. Our results show this is suboptimal: explicitly controlling the within- and between-identity variance is a first-order factor for clothes-agnostic embeddings.


Why does the intra-/inter-person balance matter?
In CC-ReID, the dominant appearance shift is structured, identity-preserving clothing change, not viewpoint or illumination. Inter-person separation alone yields embeddings discriminative for same-clothes pairs yet brittle across outfits. MINER foregrounds cross-outfit alignment, using dissimilar cross-camera outfit pairs as hard positives and visually similar identities as hard negatives. Our mining strategy ablation study on PRCC/Celeb-L shows significant gain to Rank@1 ($p_{value}<0.01$) using simultaneous hard-positive and hard-negative mining (see \cref{tab:main_sampler_ablation}).
\begin{table}[t]
\centering
\resizebox{\linewidth}{!}{
\begin{tabular}{lcc}
\toprule
\small{\textit{$\overline{R@1}$ over 3 paired seeds}} & \small{\textbf{Random-Negative}}& \small{\textbf{Hard-Negative}} \\
\midrule
\small{\textbf{Random-Positive}  }
& $67.1 / 74.3$ 
& $72.0 / 75.3$ \\
\small{\textbf{Hard-Positive}} 
& $67.5 / 76.0$ 
& $\mathbf{73.6} / \mathbf{76.4}$ \\
\bottomrule
\end{tabular}
}
\begin{flushleft}
\end{flushleft}
\vspace{-0.8em}
\caption{Sampler ablation reporting $\overline{R@1}$ on PRCC/Celeb-L. MINER uses both hard-positive and hard-negative mining.}
\label{tab:main_sampler_ablation}
\end{table}
MINER samples from soft distributions over identity and outfit centroids rather than selecting the deterministic ``hardest'' case, which is unstable under noisy hardness estimates~\cite{Wang_Hua_Kodirov_Hu_Robertson_2019}. Consistently, deterministic hard mining reduces R@1 by $3.7/0.6$ points on PRCC/Celeb-L relative to MINER. Centroid-level mining is lower-variance than per-image selection and computationally efficient~\cite{suh2019stochastic}, yielding gains in 20 finetuning epochs. Its block-diagonal supervision treats each identity as a multi-positive query--gallery set, enforcing cross-outfit invariance within blocks while discriminating against hard negatives across identities; the supplemental analysis shows this formulation is gradient-selective, concentrating optimization on confusable negatives and cross-clothing positives while suppressing redundant easy-match supervision.

Gains hold across backbones (CLIP-ReID, MADE, DIFFER, EVA-CLIP) and extend to same-clothes and RGB-Depth where modality acts as an appearance factor analogous to clothing---suggesting MINER is a general mechanism for controlling within-class variability while preserving between-class discrimination. The optimal $\frac{m}{n}$ ratio is dataset-dependent: the homogeneous, controlled PRCC favors larger $n$, while ``in-the-wild'' Celeb-L favors larger $m$. The monotonic decay of mAP on either side of the optimum indicates both extremes hurt---too few identities weakens inter-person pressure, too few samples-per-identity weakens cross-outfit alignment.




\parhead{Runtime analysis.} On a single A100 80GB GPU, MINER embeds PRCC in $54$ seconds ($17,896$ images at $329$ Hz), computes identity centroids in $13$ ms, and performs each SGD step in $517$ ms. The variance controller adds only $0.7$ ms per $T$ steps to update the batch composition $\langle m,n\rangle$, making its runtime overhead negligible.

\parhead{Implementation details.}
Experiments follow the DIFFER codebase. Images are resized to $224\times224$ and ImageNet-normalized. 
Models are trained for $20$ epochs (SGD), learning rate $2\times10^{-7}$, weight decay $0.01$, and a linear schedule with one warmup epoch, warmup ratio $0.01$, and decay factor $\gamma=0.1$. 
Batches follow $B_{\max}=174$, CLIP softmax temperature $\tau=0.1$, $\eta=0.025$, and the variance controller is updated every $T=20$ iterations.


\parhead{Limitations.}
MINER assumes vanilla-pretrained weights; training from scratch remains unexplored as unreliable early embeddings yield noisy centroids and unstable mining. It also requires clothing-ID labels and at least two outfits per identity to expose intra-person variation. Limited clothing diversity therefore weakens hard-positive mining, as in LTCC where some identities have only one clothing ID.


\parhead{Ethics.}
Person ReID can aid safety-critical retrieval but also risks privacy-invasive surveillance.
We use only public benchmarks, do not infer protected attributes, and recommend deployment only under lawful, purpose-limited governance with access control and regular privacy audits.

\parhead{Future work.}
Extending MINER from finetuning to training from scratch requires mining strategies
stable under unreliable early-stage embeddings. MINER may also apply to cross-domain metric learning tasks such as speaker identification under spoken language code-switching.

%% file: sec/8_conclussion.tex
\section{Conclusion}
\label{sec:conclusion}

{\looseness=-1
CC-ReID is challenging since clothing-induced within-identity variation differs substantially across datasets. 
We show that balancing intra- and inter-person variance during training is critical for clothes-agnostic representation learning, and introduce MINER, a simple stochastic mining framework with online closed-form batch adaptation. MINER improves retrieval across CC-ReID benchmarks and generalizes to conventional and RGB-Depth ReID, reinforcing variance-aware training as a principled path toward robust identity representation under appearance shift.
\par}

\newpage

%% file: sec/10_acknowledgements.tex
\section{Acknowledgments}
This work was supported by the Joint NSFC-ISF Research Grant number 3077/23 and ISF grant number 1574/21.

%% file: sec/supplement_sections.tex
\noindent\textbf{Overview.}
    This supplement provides reproduction details and extended analysis supporting the main
    paper. We first give the full training configuration --- backbone, normalization,
    optimization schedule, and per-dataset protocols --- across the clothes-changing,
    traditional, and RGB-Depth ReID settings, together with additional results on
    same-clothes and general ReID and on all three TVRID tracks (RGB, Depth, RGB-Depth).
    A paired-seed ablation then separates hard-positive from hard-negative mining, showing
    that hard-negative mining is the reliable component across datasets while hard-positive
    mining contributes most where intra-person variance is highest. The sampler and the
    Variance Controller act on different axes: the sampler decides which pairs enter a
    batch, the Controller how a fixed image budget is split between identities and samples
    per identity. We benchmark the latter against nine fixed allocations --- it attains the
    lowest mean and worst-case regret and the best average rank --- and show that the
    closed-form allocation risk orders measured retrieval on PRCC and Celeb-L, LTCC being a
    near-flat plateau in which no allocation is meaningfully preferred.
    We next validate the Controller's three modeling assumptions by their effect on the
    chosen allocation: the equal-weighting choice is immaterial to retrieval, the omitted
    cross-term is small and sign-favorable (so the additive risk is a conservative upper
    bound), and the projection-head proxy recovers the same allocation as the
    full-parameter gradient to within one unit per coordinate. We also show the allocation
    objective is strictly convex along the budget frontier, so its minimizer is unique.
    Finally, we analyze the contrastive objective at the gradient level, relate it to
    supervised contrastive learning, and show that bipartite query--gallery normalization
    avoids a target-mass dilution that an all-to-all formulation would incur.

\section{Implementation Details}
\label{sec:implementation_details}
Unless noted otherwise, all settings below follow the DIFFER code base, with
hyper-parameters adapted to MINER; \cref{tab:reproducibility} lists the complete
configuration. All runs use a fixed random seed ($1234$); ablations requiring repeated
measures vary only the seed.

\parhead{Backbone and inputs.}
We report results with two image encoders, DIFFER and EVA2-CLIP-L/14, both producing a
$1{,}024$-D embedding, on PRCC, LTCC, Celeb-L, DeepChange, VC-Clothes, and LaST. Inputs are
resized to $224\times224$ and normalized with the standard ImageNet statistics
$\mu=(0.485,0.456,0.406)$, $\sigma=(0.229,0.224,0.225)$.

\parhead{Optimization.}
Models are trained with SGD under a linear learning-rate schedule with a one-epoch warmup
(warmup ratio $0.01$) followed by decay with factor $\gamma=0.1$. The DIFFER-initialized
runs use $20$ epochs, learning rate $2\times10^{-7}$, and weight decay $0.01$. The
EVA-CLIP checkpoints are not pre-trained on any ReID task, so they require longer
adaptation: we finetune them for $30$ epochs at a higher learning rate of $1\times10^{-6}$,
keeping all other settings fixed.\footnote{This is the only setting that differs from
\cref{tab:reproducibility}, which otherwise reports the DIFFER-initialized configuration.}

\parhead{MINER-specific settings.}
The per-batch image budget is $2mn\le B_{\max}$ with $B_{\max}=174$, the hard memory
ceiling on an 80\,GB A100. The Variance Controller updates $\langle m,n\rangle$ every
$T=20$ iterations and smooths the gradient-variance traces with an EMA of rate
$\eta=0.025$. Soft sampling over identity and clothing centroids
(Eqs.~4 and~7, main paper) uses a softmax temperature $\tau$, following CLIP%
~\citep{radford2021learning}; $\tau=0.1$ for the single-modal datasets and $\tau=0.2$ for
the cross-modal TVRID setting (\cref{sec:tvrid}).

\begin{table}[h]
\centering
\begin{tabular}{ll}
\toprule
\textbf{Hyper-parameter} & \textbf{Value} \\
\midrule
Input image size & $224 \times 224$ \\
Normalization mean ($\mu$) & $(0.485, 0.456, 0.406)$ \\
Normalization std.\ ($\sigma$) & $(0.229, 0.224, 0.225)$ \\
\midrule
Optimizer & SGD \\
Learning rate & $2 \times 10^{-7}$ \\
LR scheduler & Linear \\
Weight decay & $0.01$ \\
Total epochs & $20$ \\
Warmup epochs & $1$ \\
Warmup ratio & $0.01$ \\
LR decay factor ($\gamma$) & $0.1$ \\
Random seed & $1234$ \\
\midrule
Batch budget, $B_{\max}$ & $174$ \\
Controller interval ($T$) & $20$ \\
Controller EMA rate ($\eta$) & $0.025$ \\
Softmax temperature ($\tau$) & $0.1$ \\
\bottomrule
\end{tabular}
\caption{Default training configuration (DIFFER-initialized runs). $B_{\max}$ is the hard
memory ceiling on the per-batch image budget $2mn$. EVA-CLIP-initialized runs use $30$
epochs at learning rate $1\times10^{-6}$; the cross-modal TVRID setting uses $\tau=0.2$
(\cref{sec:tvrid}).}
\label{tab:reproducibility}
\end{table}

\section{Experiments}
We follow each benchmark's official evaluation protocol and the conventions of its original
work. Throughout, retrieval is single-query, each query is scored against the gallery
without re-ranking, and no test-set statistic or prior is used; results are reported as
Rank@1 accuracy (R@1) and mean Average Precision (mAP). The per-dataset protocols are: PRCC
on the cross-clothes Camera~A$\rightarrow$C split, LTCC on its official clothes-changing
split, Celeb-L on its inherently clothes-changing query/gallery split, DeepChange on its
long-term probe/gallery protocol, VC-Clothes on the Camera~3/4 clothes-changing split, and
LaST on its large-scale spatio-temporal protocol.

\subsection{Clothes-Changing ReID}
For the six CC-ReID datasets (PRCC, LTCC, Celeb-ReID-Light, DeepChange, VC-Clothes, LaST) we
evaluate under the cross-clothing setting, in which query and gallery images show different
outfits of the same identity wherever the protocol defines such pairs. The corresponding
results appear in Tab.~1 of the main paper.

\subsection{Traditional ReID}
While MINER targets the clothes-changing setting, conventional fixed-clothing retrieval
remains central to deployment, so we additionally evaluate two same-clothing protocols:
PRCC in the \emph{Same-Clothes} (SC) protocol and LTCC in the \emph{General} protocol
(\cref{tab:sc_reid}). For comparability we match the backbone and evaluation settings of the
prior work we compare against.

On LTCC-General, MINER sets a new state of the art by a wide margin
($93.8$/$67.7$ vs.\ the next-best $85.0$/$52.8$ in R@1/mAP), despite prioritizing identity
over clothing cues. PRCC-SC is effectively saturated --- seven methods reach $100.0$ R@1 ---
so it offers little discriminative headroom; there MINER is competitive on mAP ($99.4$)
without surpassing the top entries. The strong LTCC-General result is the more informative
of the two: it shows that MINER's clothes-agnostic training does not sacrifice same-clothes
retrieval, and in fact improves it where the benchmark is not already maxed out.

\begin{table}[t]
\begin{center}
\begin{small}
\centering
\begin{tabular}{lcccc}
\toprule
\textbf{\small{SC-ReID}} & \multicolumn{2}{c}{\textbf{\small{PRCC}}}  & \multicolumn{2}{c}{\textbf{\small{LTCC}}}  \\
\textbf{\small{Method}} & \small{R@1} & \small{mAP}& \small{R@1} & \small{mAP} \\
\midrule
\small{TransReID} \tiny{`21}   & 97.3 & 95.9 & 73.8 & 39.4 \\
\small{CAL} \tiny{`22}   & \textbf{100.0} & 99.8 & 74.2 & 40.8 \\
\small{MADE} \tiny{`24}   & \textbf{100.0} & 98.6 & 84.2 & 48.2 \\
\small{RFLNet} \tiny{`25}   & \textbf{100.0} & 99.6 & -- & -- \\
\small{FA-ReID} \tiny{`25}   & \textbf{100.0} & 99.0 & 75.7 & 41.3 \\
\small{CSCI} \tiny{`25}   & \textbf{100.0} & \textbf{99.9} & 82.6 & 48.0 \\
\small{DIFFER} \tiny{`25} & \textbf{100.0} & 98.6 & \underline{85.0} & \underline{52.8} \\
\midrule
\small{MINER \tiny{(Ours)}} & \textbf{100.0} & \underline{99.4} & \textbf{93.8} & \textbf{67.7} \\
\bottomrule
\end{tabular}
\end{small}
\end{center}
\caption{Same-clothes (PRCC-SC) and General (LTCC) ReID vs.\ prior work, in R@1/mAP. Best is
\textbf{bold}, second \underline{underlined}. PRCC-SC R@1 is saturated at $100.0$ across
methods; MINER sets a new state of the art on LTCC-General.}
\label{tab:sc_reid}
\end{table}

\subsection{MINER as a Finetuning Framework}
\subsubsection{Backbone Gains}
Table~2 of the main paper reports the gain MINER delivers on top of existing backbones.
For each backbone we reproduce the original training procedure to convergence, record the
reproduced base scores, and then apply MINER as a finetuning stage on the converged model.
Our reproduced base numbers occasionally differ from the originally reported ones, as is
common when re-training across code bases and hardware. This does not affect the conclusion:
the quantity of interest is the \emph{delta} from adding MINER, and measuring base and
$+$MINER under identical settings on our own reproduction isolates that delta more cleanly
than a cross-paper comparison would. The consistent improvement across backbones therefore
quantifies MINER's value as a generic, architecture-agnostic finetuning step.

\subsubsection{Baseline}
The EVA-CLIP backbone is a strong general purpose image encoder. We further report its expressive power for the CC-ReID task both as a zero-shot (off-the-shelf) vanilla encoder and as a baseline with a simple approach like multi-class classifier finetuning using the simple ID-Loss.
This analysis was conducted by \citet{liang2025differ} on EVA-CLIP and demonstrated in \cref{tab:eva_baseline} over the PRCC, LTCC, and Celeb-L datasets.
MINER introduces net mAP gains of 46.1, 29.3, and 22.0 respectively compared with the vanilla EVA-CLIP and 7.0, 4.0, and 5.8 over the baseline.

\begin{table}[t]
\centering
\resizebox{\linewidth}{!}{
\begin{tabular}{lcc|cc|cc}
\toprule

\multirow{2}{*}{Method}
& \multicolumn{2}{c|}{PRCC}
& \multicolumn{2}{c|}{LTCC}
& \multicolumn{2}{c}{Celeb-L} \\
& R@1 & mAP
& R@1 & mAP
& R@1 & mAP \\
\midrule
EVA-CLIP (No finetune)
& 39.3 & 23.9
& 15.0 & 5.1
& 57.7 & 37.1 \\
Baseline (ID Loss)
& 65.1 & 63.0
& 54.6 & 30.4
& 72.7 & 53.3 \\
DIFFER
& 68.5 & 64.7
& 58.2 & 31.6
& 75.6 & 54.3 \\
\midrule
MINER (EVA-CLIP)
& \textbf{74.5} & \textbf{70.0}
& \underline{60.2} & \textbf{34.4}
& \textbf{77.2} & \textbf{59.1} \\
MINER (DIFFER)
& \underline{73.9} & \underline{69.7}
& \textbf{60.9} & \underline{33.3}
& \underline{76.9} & \underline{56.3} \\

\bottomrule
\end{tabular}
}
\caption{MINER versus three initialization baselines on PRCC, LTCC, and Celeb-L
under the clothes-changing protocol, in R@1/mAP. Upper block: EVA-CLIP off the shelf,
EVA-CLIP finetuned with a plain ID-loss, and the ReID-pretrained
DIFFER encoder. Lower block: MINER applied
to each of the two initializations. Best per column in \textbf{bold}, second
\underline{underlined}.}
\label{tab:eva_baseline}
\end{table}

\subsection{TVRID 2026 Competition}
\label{sec:tvrid}
The ICPR 2026 Competition on Privacy-Preserving Person Re-Identification from Top-View
RGB-Depth Camera (TVRID)~\cite{delecluse2026icpr} drew $732$ submissions from $148$ research
groups across three tracks (RGB, Depth, RGB-Depth). As \cref{tab:top3_and_average_per_task}
shows, MINER is the only method to place in the top three of \emph{all} three tracks; every
other leading method reaches the podium in at most two. The wide gap between the podium
scores and the per-track averages (e.g.\ $94.9$ vs.\ $40.7$ mAP on RGB-Depth) indicates the
setting is far from saturated --- top-view sensors hide the face, so identity must be
recovered from body shape and gait alone.

\begin{figure*}[t]
\centering
\includegraphics[width=1\linewidth,trim=0.15cm 1.35cm 1cm 0cm,clip]{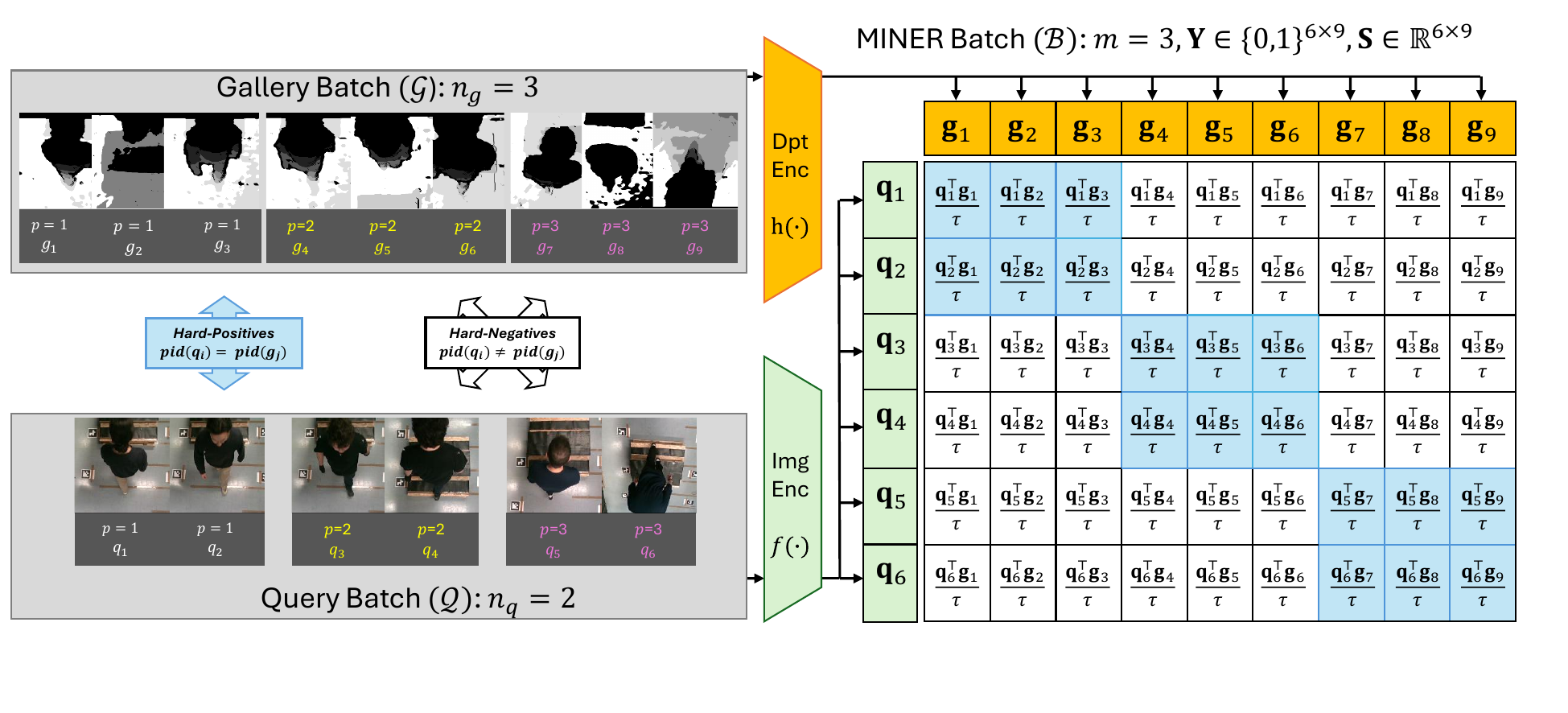}
\caption{MINER's contrastive loss over a cross-modal bipartite batch. Hard-negative
identities and hard-positive cross-modal pairs distill inter-person (white) and intra-person
(blue) variance simultaneously via a block-diagonal label matrix, here over the RGB-Depth
data of the TVRID 2026 Competition.}
\label{fig:teaser_tvrid}
\end{figure*}

\parhead{Cross-modal adaptation.}
The RGB-Depth track maps a query in one modality to a gallery in another, so we adapt
MINER's appearance factor from clothing to modality: hard positives are now cross-modal
pairs of the same identity and hard negatives are same-modality look-alikes, controlling
intra- and inter-person variance exactly as in the single-modal case
(\cref{fig:teaser_tvrid}). We instantiate two EVA-CLIP-L/14 encoders from the DIFFER
checkpoints, one per modality, and finetune on the TVRID data.

Two changes accommodate the cross-modal setting. First, query and gallery are no longer
interchangeable, which breaks the symmetry between $\mathcal{L}_{q\rightarrow g}$ and
$\mathcal{L}_{g\rightarrow q}$; we therefore decouple the per-side image counts, allowing
$n_q\neq n_g$ and modeling the budget as $B=m(n_q+n_g)$ rather than the single-modal
$n_q=n_g=n$. This extends the allocation knob to the $3$-tuple $\langle m,n_q,n_g\rangle$.
Second, the larger RGB$\leftrightarrow$Depth domain shift produces a flatter similarity
distribution, so we raise the softmax temperature to $\tau=0.2$ (from $0.1$) to keep the
contrastive targets well-scaled. All other settings follow
\cref{sec:implementation_details}.

\begin{table}[t]
\centering
\resizebox{\linewidth}{!}{
\begin{tabular}{lllcc}
\toprule
\textbf{Task} & \textbf{Rank} & \textbf{Method} & \small{R@1} & \small{mAP} \\
\midrule
\multirow{4}{*}{RGB}
& 1st  & Duy \textit{et al.}          & 100.0 & 100.0 \\
& 2nd  & Carlo Metta                  & 100.0 & 99.9  \\
& 3rd  & MINER (Ours)                 & 99.8  & 99.9  \\
& Avg. & \textit{Task average}        & 85.7  & 89.6  \\
\midrule
\multirow{4}{*}{Depth}
& 1st  & Jiang \textit{et al.}        & 98.8  & 99.4  \\
& 2nd  & MINER (Ours)                 & 97.2  & 98.3  \\
& 3rd  & Rashidunnabi \textit{et al.} & 90.4  & 94.3  \\
& Avg. & \textit{Task average}        & 47.1  & 57.2  \\
\midrule
\multirow{4}{*}{RGB-D}
& 1st  & Rashidunnabi \textit{et al.} & 99.8  & 99.9  \\
& 2nd  & Jiang \textit{et al.}        & 97.9  & 98.7  \\
& 3rd  & MINER (Ours)                 & 91.5  & 94.9  \\
& Avg. & \textit{Task average}        & 30.4  & 40.7  \\
\bottomrule
\end{tabular}
}
\caption{TVRID top-3 entries and per-track average over $732$ submissions. MINER is the only
method on the podium of all three tracks. The large podium-to-average gap, especially on
Depth and RGB-Depth, reflects an unsaturated, face-free setting.}
\label{tab:top3_and_average_per_task}
\end{table}

\subsection{MINER's Sampler: Ablation Study}
The main thesis of MINER rests on the interaction of two factors: the intra-class and inter-class variance effects. To quantify their individual contributions, we report the paired $t$-test analysis in \cref{tab:sampler_ablation_prcc_celebl}. Seeds were shared across configurations, yielding a paired design with $\mathrm{dof}=2$.
The full MINER sampler (hard-positive and hard-negative combined) produces the largest and most significant Rank@1 gain on both datasets: $+6.50$ ($p_{\mathrm{value}}=0.0007$) on PRCC and $+2.17$ ($p_{\mathrm{value}}=0.0059$) on Celeb-L.
Decomposing the two factors reveals a dataset-dependent pattern that is consistent with our variance analysis. Hard-negative mining is the reliable component: it improves significantly over random mining on \emph{both} datasets ($+4.87$, $p_{\mathrm{value}}=0.0002$ on PRCC; $+1.03$, $p_{\mathrm{value}}=0.0072$ on Celeb-L). Hard-positive mining alone, in contrast, is not significant on PRCC ($+0.33$, $p_{\mathrm{value}}=0.29$) and only marginal on Celeb-L ($+1.73$, $p_{\mathrm{value}}=0.052$), yet its point estimate is largest precisely on Celeb-L.
Complementarily, the Variance Controller independently adapts $m:n$ according to the measured gradient-variance ratio. Thus, the sampler ablation reveals where cross-clothing positives are most beneficial, while the controller determines how the training budget should be reallocated.
\begin{table*}[t]
\centering

\small
\resizebox{\linewidth}{!}{
\begin{tabular}{llccc}
\toprule
\textbf{Dataset} & \textbf{Setting/Factor} & \textbf{Sampler configuration} 
& \textbf{$\overline{\text{R@1}}$ (\%)}
& \textbf{$\Delta$R@1, ~~$p_{value}$} 
\\
\midrule
PRCC 
& (i)   
& Random positive + random negative       
& $67.13 \pm 0.25$ 
& -- \\

& (ii)  
& Random positive + hard negative         
& $72.00 \pm 0.26$ 
& $+4.87$, $p_{value}=0.0002$ 
\\

& (iii) 
& Hard positive + random negative         
& $67.47 \pm 0.25$ 
& $+0.33$, $p_{value}=0.29$ 
\\

& (iv)  
& MINER: hard positive + hard negative    
& $\mathbf{73.63 \pm 0.30}$ 
& $\mathbf{+6.50}$, $p_{value}=0.0007$ 
\\

\midrule
Celeb-L 
& (i)   
& Random positive + random negative       
& $74.27 \pm 0.60$ 
& -- \\

& (ii)  
& Random positive + hard negative         
& $75.30 \pm 0.53$ 
& $+1.03$, $p_{value}=0.0072$ 
\\

& (iii) 
& Hard positive + random negative         
& $76.00 \pm 0.20$ 
& $+1.73$, $p_{value}=0.0516$ 
\\

& (iv)  
& MINER: hard positive + hard negative    
& $\mathbf{76.43 \pm 0.40}$ 
& $\mathbf{+2.17}$, $p_{value}=0.0059$ 
\\
\bottomrule
\end{tabular}
}
\caption{Sampler-mining ablation on PRCC and Celeb-ReID-Light over three paired seeds. Results report mean $\pm$ standard deviation over 3 seeds. Paired $t$-tests compare each configuration against the random-positive/random-negative baseline (i), with $\mathrm{dof}=2$.}
\label{tab:sampler_ablation_prcc_celebl}
\end{table*}


\begin{table}[t]
\centering
\small
\begin{tabular}{lcc}
\toprule
\textbf{Dataset} & \textbf{Deterministic Hard Mining} & \textbf{MINER} \\
& \textbf{$\overline{\text{R@1}}$ (\%)} & \textbf{$\overline{\text{R@1}}$ (\%)} \\
\midrule
PRCC
& $69.93 \pm 0.15$
& $\mathbf{73.63 \pm 0.30}$ \\

Celeb-L
& $75.87 \pm 0.09$
& $\mathbf{76.43 \pm 0.40}$ \\
\bottomrule
\end{tabular}

\caption{Deterministic mining selects the hardest negative identities and hardest positive clothing pairs, while MINER samples stochastically from the corresponding hardness distributions. Results report mean $\pm$ sample standard deviation over 3 seeds. MINER consistently improves R@1 over deterministic hard mining, with a larger gain on PRCC than on Celeb-L.}
\label{tab:deterministic_sampler_ablation}
\end{table}

While random mining provides a natural lower-bound baseline, we additionally evaluate the opposite extreme: deterministic hard mining. Specifically, instead of sampling from MINER's soft hardness distributions, we deterministically select the $m$ most similar negative identities and, for each identity, the most dissimilar cross-clothing positive pair. Using the same paired seeds as in \cref{tab:sampler_ablation_prcc_celebl}, this experiment isolates the contribution of MINER's stochastic sampling from that of hard-example selection itself. As shown in \cref{tab:deterministic_sampler_ablation}, deterministic hard mining reaches $\overline{\mathrm{R@1}}=69.93\pm0.15$ on PRCC and $75.87\pm0.09$ on Celeb-L, compared with $73.63\pm0.30$ and $76.43\pm0.40$ for MINER, respectively. Thus, deterministic selection improves substantially over random mining, confirming the value of hard-example selection, but remains below stochastic MINER by $3.70$ R@1 points on PRCC and $0.56$ points on Celeb-L. The consistent advantage in R@1 supports MINER's use of soft hard mining: biasing training toward difficult examples while retaining stochasticity is preferable to repeatedly selecting only the most extreme samples. The smaller gap on Celeb-L further indicates that the benefit of stochasticity is dataset-dependent rather than uniform.

\subsection{Multi-Positive Objective and Variance Controller: Ablation Study}
To isolate MINER's objective and adaptive batch allocation from its hard-mining sampler, we combine the controlled comparisons of \cref{tab:eva_baseline} and \cref{tab:sampler_ablation_prcc_celebl}. Specifically, the random-positive/random-negative setting disables hard mining while retaining MINER's block-diagonal multi-positive objective and Variance Controller. Compared with the matched ID-loss baseline, this configuration improves $\overline{\mathrm{R@1}}$ by $+2.0$ points on PRCC (from $65.1\%$ to $67.1\%$) and $+1.6$ points on Celeb-L (from $72.7\%$ to $74.3\%$). These gains show that MINER's improvement is not solely attributable to hard-example mining; the multi-positive objective and adaptive allocation provide an independent contribution.

\subsection{MINER's Variance Controller: Ablation Study}
\label{sec:alloc_stats}
We compare the Variance Controller against the nine fixed allocations on the
PRCC/LTCC/Celeb-L grids (\cref{tab:regret_supp}). Because mAP scales differ across
datasets, we analyze \emph{regret}---the gap, in mAP points, to each dataset's grid
optimum---so that the per-dataset oracle has regret $0$ and every policy is measured on
a common scale.

\parhead{Minimax regret (primary).}
At training time the dataset's intra-/inter-person prior is unknown, so a fixed
allocation is chosen blind and its quality is governed by its worst case across
datasets~\cite{savage1951}. The controller attains a worst-case regret of $0.2$ mAP and
a mean of $0.10$, against a best fixed allocation of $0.6$ worst-case ($\langle 7,12\rangle$)
and $0.33$ mean ($\langle 5,17\rangle$)---attained, notably, by \emph{different} splits.
The controller's mean and worst-case regret are below those of \emph{every} fixed
allocation (\cref{tab:regret_supp}, last two columns), so it Pareto-dominates all
nine on both criteria simultaneously, improving the best fixed choice by $3.3\times$ in
mean and $3\times$ in worst-case regret. This is a property of the measured grid and
requires no distributional assumption.

\parhead{Average rank (companion).}
As a standard cross-dataset summary~\cite{demsar2006}, we rank the ten configurations
within each dataset and average. The controller obtains the best mean rank, $3.33$,
ahead of every fixed split (next best $4.50$). Tellingly, the three aggregate criteria
above each favor a different fixed allocation, yet the controller leads on all three---no
fixed split is robustly best, whereas the controller is.

\parhead{On significance testing.}
An omnibus Friedman test over the three datasets is not significant
($\chi^2_F(9)\!=\!6.6$), and is necessarily underpowered at $N\!=\!3$: the Bonferroni--Dunn
critical difference ($\approx\!6.9$ ranks) exceeds the entire observed rank range
($5.5$), so no post-hoc comparison can clear it regardless of effect size. We therefore
report these gaps descriptively rather than as inferential claims. Statistical power
here resides not in the three datasets but in the seed replicates of the main results
(Sec.~5): the controller's largest shortfall ($0.2$ mAP) lies within
their run-to-run variation, while the transfer penalties of fixed splits
($\langle 5,17\rangle$: $0.9$ mAP on Celeb-L; $\langle 21,4\rangle$: $2.1$ on PRCC) exceed
it severalfold. The regret gap can only widen as datasets with more varied priors are added.
\begin{table*}[t]
\centering
\small
\setlength{\tabcolsep}{4pt}
\begin{tabular}{l ccc ccc cc c}
\toprule
& \multicolumn{3}{c}{mAP} & \multicolumn{3}{c}{Regret} & \multicolumn{2}{c}{Agg.\ regret} & \\
\cmidrule(lr){2-4}\cmidrule(lr){5-7}\cmidrule(lr){8-9}
$\langle m,n\rangle$ & PRCC & LTCC & Celeb-L & PRCC & LTCC & Celeb-L & Mean & Worst & Rank \\
\midrule
$\langle 29,3\rangle$ & 67.6 & 33.9 & 55.7 & 2.1 & 0.6 & 0.8 & 1.17 & 2.1 & 8.83 \\
$\langle 21,4\rangle$ & 67.6 & 34.0 & \underline{56.5} & 2.1 & 0.5 & 0.0 & 0.87 & 2.1 & 6.50 \\
$\langle 17,5\rangle$ & 68.2 & 34.4 & 56.3 & 1.5 & 0.1 & 0.2 & 0.60 & 1.5 & 5.50 \\
$\langle 12,7\rangle$ & 68.6 & 34.2 & 56.4 & 1.1 & 0.3 & 0.1 & 0.50 & 1.1 & 5.17 \\
$\langle 9,~~9\rangle$  & 68.9 & 34.4 & 56.3 & 0.8 & 0.1 & 0.2 & 0.37 & 0.8 & 4.50 \\
$\langle 7,12\rangle$ & 69.1 & 34.1 & 56.2 & 0.6 & 0.4 & 0.3 & 0.43 & \underline{0.6} & 6.00 \\
$\langle 5,17\rangle$ & \underline{69.7} & 34.4 & 55.6 & 0.0 & 0.1 & 0.9 & \underline{0.33} & 0.9 & 4.67 \\
$\langle 4,21\rangle$ & 69.6 & \underline{34.5} & 55.5 & 0.1 & 0.0 & 1.0 & 0.37 & 1.0 & 4.50 \\
$\langle 3,29\rangle$ & 68.6 & \underline{34.5} & 55.1 & 1.1 & 0.0 & 1.4 & 0.83 & 1.4 & 6.00 \\
\midrule
MINER Controller & 69.7 & 34.4 & 56.3 & 0.0 & 0.1 & 0.2 & \textbf{0.10} & \textbf{0.2} & \textbf{3.33} \\
\bottomrule
\end{tabular}
\caption{Per-allocation mAP and regret (gap to each dataset's grid optimum) on
PRCC/LTCC/Celeb-L, with mean and worst-case regret and Dem\v{s}ar average rank across the
three. The Variance Controller attains the lowest mean and worst-case regret of all
configurations---Pareto-dominating every fixed split---and the best average rank. Lower
is better throughout; per-dataset optima are underlined.}
\label{tab:regret_supp}
\end{table*}

\begin{figure*}[t]
    \centering
    \includegraphics[width=\textwidth]
    {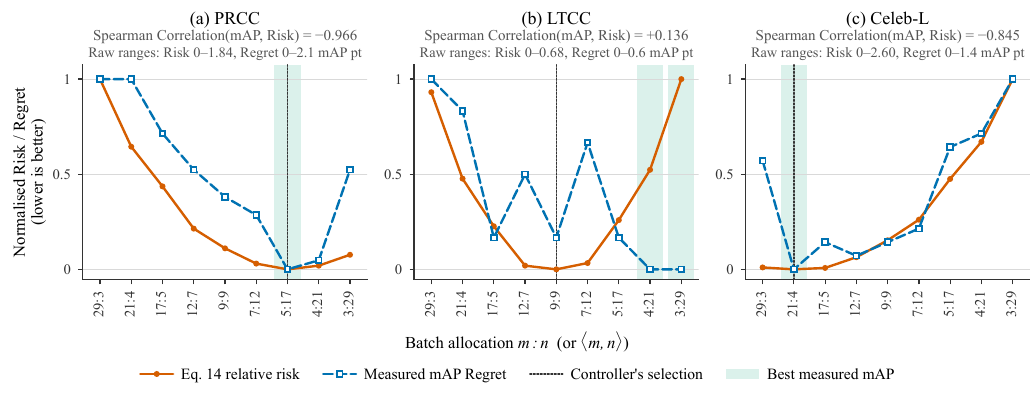}
    \caption{
    Risk (orange) versus mAP regret (blue) across Tab.~\ref{tab:regret_supp}'s nine
    allocations, each normalized within its dataset; lower is better. Dashed lines mark
    the Controller's choice --- the risk minimum by construction (Eq.~15) --- and shaded
    cells the best measured. The test is whether the curves track. They do on PRCC
    ($r_s=-0.97$) and Celeb-L ($r_s=-0.85$), where lowest risk is also best measured.
    LTCC spans only $0.6$ mAP, too flat to order ($r_s=+0.14$, $p_{value}=0.73$); the
    Controller's $\langle9,9\rangle$ lands $0.1$ mAP off.
    }
    \label{fig:eq14_vs_map}
\end{figure*}

\parhead{Visualizing the effect.}
Tab.~\ref{tab:regret_supp} reports the measured mAP and regret for every fixed
allocation. Fig.~\ref{fig:eq14_vs_map} plots the same grid against the allocation
risk of Eq.~14, evaluated per dataset at its estimated controller ratio
$\rho=m^\star/n^\star$; only relative risk is needed, so the common covariance scale
cancels, and risk and regret are independently normalized to $[0,1]$ within each
dataset for display.
Two comparisons appear in the figure and they carry different weight. That the risk
is minimized \emph{at} the Controller's selection is not evidence: by Eq.~15 the
Controller returns the argmin of Eq.~14, so the dashed line and the orange minimum
coincide by construction. The substantive question is whether that risk tracks
\emph{measured} retrieval --- a quantity the grid supplies independently of the
Controller.
It does, wherever the grid can resolve it. The Spearman rank correlation between
risk and mAP is $r_s=-0.97$ on PRCC ($p_{value}<0.001$) and $r_s=-0.85$ on Celeb-L
($p_{value}=0.004$): lower predicted risk accompanies higher measured mAP, and on both
datasets the risk-minimizing cell is also the best-measured cell
($\langle5,17\rangle$ and $\langle21,4\rangle$). LTCC is uninformative rather than
contradictory --- its nine allocations span only $0.6$ mAP in total, comparable to
run-to-run variation, so their ordering is largely noise and the correlation is
correspondingly small and not significant ($r_s=+0.14$, $p_{value}=0.73$). Consistent with a
flat objective, the Controller's $\langle9,9\rangle$ selection there sits $0.1$ mAP
below the best measured cells.

\subsection{Variance Controller Assumption Validation}
\label{sec:assumption_validation}
The adaptive allocation of Alg.~2 (main paper, Sec.~3.3) rests on three
modeling choices. Each affects \emph{only} the selection of $\langle m,n\rangle$, never the
contrastive objective itself:
\emph{(i)} the allocation objective is the equally weighted sum of the two estimation risks;
\emph{(ii)} the between- and within-identity risks decouple, so $\mathcal{E}$ is additive; and
\emph{(iii)} projection-head gradients are a low-variance proxy for the full-parameter
covariance traces. We validate all three by their effect on the chosen allocation, since
that --- not the intermediate quantities --- is what the controller produces.

\begin{figure}[t]
    \centering
    \includegraphics[width=\linewidth]{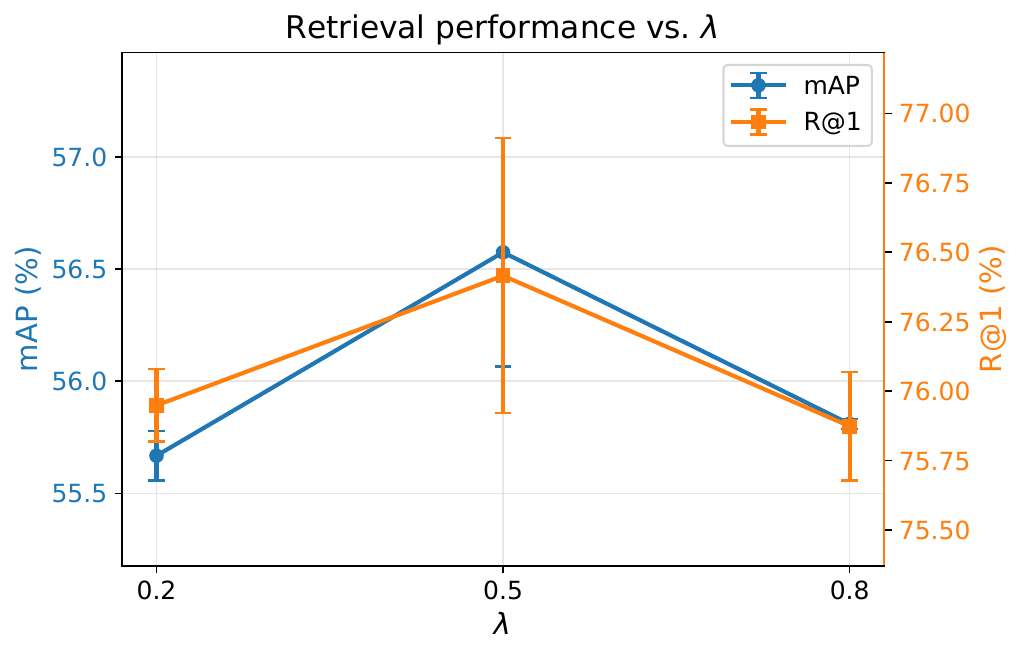}
    \includegraphics[width=\linewidth]{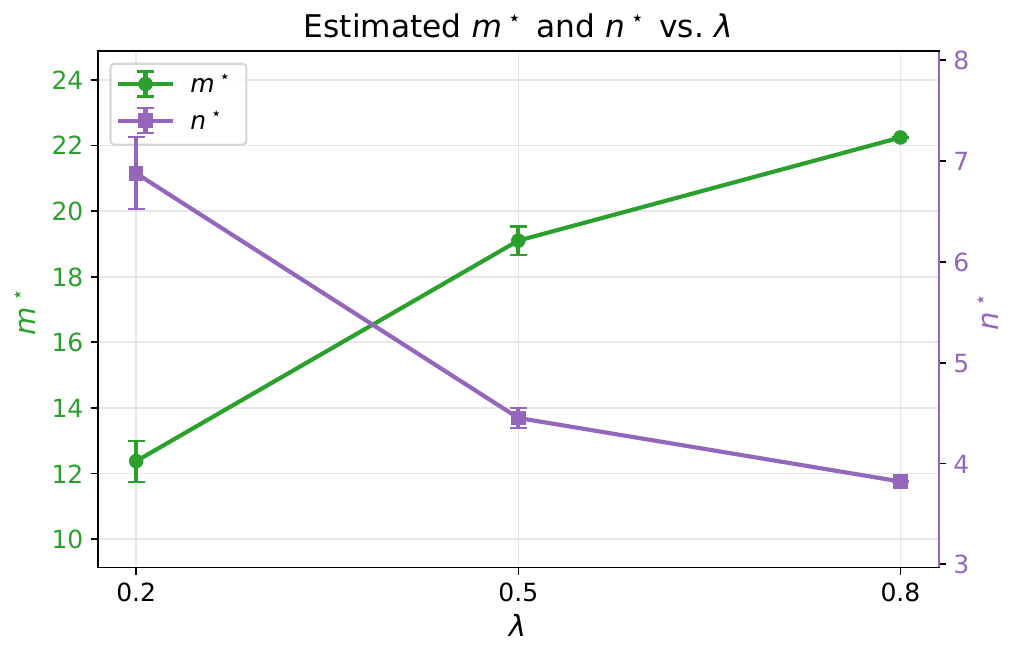}
    \caption{Effect of the risk weighting $\lambda$ on Celeb-L.
    \textbf{Top:} retrieval performance (mAP, R@1) is flat across $\lambda$.
    \textbf{Bottom:} the controller allocation $\langle m^\star,n^\star\rangle$ shifts
    markedly with $\lambda$. The allocation moves while performance does not, confirming the
    equal-weighting choice is immaterial. Mean over three seeds; error bars are standard
    deviation.}
    \label{fig:lambda_sensitivity}
\end{figure}

\parhead{Assumption (i): equal weighting.}
The $1/m$ and $1/2n$ scalings in $\mathcal{E}$ are fixed by the corresponding estimator
variances; the only free choice is the relative weight on the two terms, which we set to
$1{:}1$. This choice is not load-bearing. Generalizing to
\begin{equation*}
\mathcal{E}_\lambda(m,n)=\lambda \frac{\operatorname{tr}\Sigma_{\mathrm{B}}}{m}
+(1-\lambda) \frac{\operatorname{tr}\Sigma_{\mathrm{W}}}{2n},
\quad \text{s.t.}\quad 2mn = B_{\max},
\end{equation*}
the constrained minimizer satisfies
\begin{equation*}
\rho_\lambda \;=\; \frac{m^\star}{n^\star}
\;=\;\frac{\lambda}{1-\lambda}\cdot\frac{2\operatorname{tr}\Sigma_{\mathrm{B}}}{\operatorname{tr}\Sigma_{\mathrm{W}}},
\end{equation*}
so $\lambda$ enters only through the scalar multiplier $\lambda/(1-\lambda)$: it slides the
operating point along the budget frontier but cannot change the functional form, recovering Eq.~15 at $\lambda=0.5$. 
Since mAP is a smooth, single-peaked function of the allocation
(main paper, Figs.~2--3), a bounded reweighting of $\rho$ moves the operating point a bounded
distance along a flat ridge. \cref{fig:lambda_sensitivity} confirms this on Celeb-L: as
$\lambda$ varies over $\{0.2,0.5,0.8\}$ the estimated allocation $\langle m^\star,n^\star\rangle$
shifts substantially, yet R@1 and mAP vary within seed noise, with $\lambda=0.5$ at least as
good as either alternative. The equal-weighting choice therefore does not affect retrieval.

\parhead{Assumption (ii): risk decoupling.}
The additive form $\mathcal{E}(m,n)=\operatorname{tr}\Sigma_{\mathrm{B}}/m+\operatorname{tr}\Sigma_{\mathrm{W}}/2n$
(Eq.~14) has two sub-claims: \emph{(ii-a) attribution} --- the between-identity dispersion
$\operatorname{tr}\Sigma_{\mathrm{B}}$ is carried mainly by the discrimination signal
$h_{\mathrm{disc}}$ (off-block / negatives) and the within-identity dispersion
$\operatorname{tr}\Sigma_{\mathrm{W}}$ by the invariance signal $h_{\mathrm{inv}}$
(within-block / positives), Eq.~12; and \emph{(ii-b) decoupling} --- their cross-covariance
is small enough that no cross-term survives.

We stress that the deployed estimators (Eqs.~16--17) recover $\operatorname{tr}\Sigma_{\mathrm{B}}$
and $\operatorname{tr}\Sigma_{\mathrm{W}}$ directly from the \emph{full} per-sample gradient by
one-way ANOVA; they do not use the $h_{\mathrm{inv}}/h_{\mathrm{disc}}$ split and do not assume
decoupling. The split motivates \emph{why} the two terms respond to $m$ and $n$; the assumption
governs only whether $\mathcal{E}$ may omit the cross-term. We assess this two ways.

First, the cross-term has a favorable sign. Writing $g_{\mathrm{inv}}(x),g_{\mathrm{disc}}(x)$
for the per-sample contributions to $h_{\mathrm{inv}},h_{\mathrm{disc}}$, their trace-correlation
\begin{equation*}
 r_{\mathrm{cross}}
 \;=\;
 \frac{\operatorname{tr}\mathrm{Cov}\!\big(g_{\mathrm{inv}},g_{\mathrm{disc}}\big)}
      {\sqrt{\operatorname{tr}\mathrm{Cov}\!\big(g_{\mathrm{inv}}\big)\,
             \operatorname{tr}\mathrm{Cov}\!\big(g_{\mathrm{disc}}\big)}}\,,
 ~~ r_{\mathrm{cross}}\in[-1,1],
\end{equation*}
is not large in magnitude both in PRCC and LTCC (\cref{tab:controller_estimator_agreement}). By the identity
$\operatorname{tr}\mathrm{Cov}(g_{\mathrm{inv}}{+}g_{\mathrm{disc}})
=\operatorname{tr}\mathrm{Cov}(g_{\mathrm{inv}})+\operatorname{tr}\mathrm{Cov}(g_{\mathrm{disc}})
+2\operatorname{tr}\mathrm{Cov}(g_{\mathrm{inv}},g_{\mathrm{disc}})$, a negative cross-term makes
the true coupled risk \emph{smaller} than the additive $\mathcal{E}$. Dropping it therefore
yields a conservative upper bound on the risk, not an uncontrolled approximation.

Second, and decisively, the omission does not change the output: the allocation obtained
under the additive surrogate matches the per-dataset grid optimum
(\cref{sec:alloc_stats}) and the full-parameter solution
(\cref{tab:controller_estimator_agreement}). Whatever residual coupling exists, it does not
move $\langle m,n\rangle$. Assumption (ii) is thus an approximation that is both conservative
and allocation-preserving.

\parhead{Assumption (iii): projection-head proxy.}
For efficiency the controller evaluates the traces on the projection head, $\nabla_{\mathbf{z}}\ell$,
rather than the full-parameter gradient $\nabla_\theta\ell$ --- the standard low-variance
gradient surrogate of~\citet{mccandlish2018empirical}. Only the \emph{ratio}
$\rho=2\operatorname{tr}\Sigma_{\mathrm{B}}/\operatorname{tr}\Sigma_{\mathrm{W}}$ enters Eq.~15,
and a ratio is robust to the (large, roughly multiplicative) magnitude gap between head and
full gradients, which cancels in the quotient. We verify this where it matters --- in allocation
space, with integer tolerance --- rather than by comparing raw traces. As
\cref{tab:controller_estimator_agreement} shows, the projection-head and full-parameter
estimators yield the same $\langle m,n\rangle$ on PRCC (exact) and agree within one unit per
coordinate on LTCC. The cheap surrogate reproduces the allocation of the full gradient.

\begin{table}[t]
\centering
\small
\begin{tabular}{lccc}
\toprule
& \multicolumn{2}{c}{\textbf{Allocation $\langle m,n\rangle$ (iii)}} & \textbf{(ii)} \\
\cmidrule(lr){2-3}\cmidrule(lr){4-4}
\textbf{Dataset}
& \textbf{Proj.\ $\nabla_{\mathbf{z}}\ell$}
& \textbf{Full $\nabla_\theta\ell$}
& \textbf{$r_{\mathrm{cross}}$} \\
\midrule
PRCC      & $\langle 5,17\rangle$ & $\langle 5,17\rangle$ \;(exact)      & $-0.23$ \\
LTCC      & $\langle 9,~~9\rangle$  & $\langle 10,8\rangle$ \;(within $1$) & $-0.25$ \\
\bottomrule
\end{tabular}
\caption{Validation of Assumptions (ii) and (iii). The last column reports the
trace-correlation $r_{\mathrm{cross}}$ between the invariance and discrimination gradient
signals (Assumption ii); it is small and negative, so the additive risk is a conservative
upper bound. The first two columns report the allocation recovered by the projection-head
versus the full-parameter estimator (Assumption iii); they agree, exactly on PRCC and within
one unit per coordinate on LTCC.}
\label{tab:controller_estimator_agreement}
\end{table}

\section{Extended Theoretical Analysis}
In this section we first detail the per-sample gradient definition and then dive into the deeper meaning of our loss function.

\subsection{Per-Query Gradient and the Two-Signal Split}
\label{app:grad}
This appendix derives the gradient decomposition of our proposed block-diagonal loss function.
With $p_{ij} = \frac{\exp(S_{ij})}{\sum_k\exp(S_{ik})}$ and normalized target
$\bar{y}_{ij}=\frac{Y_{ij}}{n}$, the per-query loss is
$\ell_i = -\tfrac{1}{n}\sum_j Y_{ij}\log p_{ij}$. 
Using the softmax identity
$\frac{\partial\log p_{ij}}{\partial S_{ik}}=\delta_{jk}-p_{ik}$ and $\sum_j Y_{ij}=n$,
\begin{align}
\frac{\partial \ell_i}{\partial S_{ik}}
&= -\frac{1}{n}\sum_j Y_{ij}\big(\delta_{jk}-p_{ik}\big) \notag \\
&= p_{ik}-\bar{y}_{ik}
=
\begin{cases}
p_{ik}-\tfrac{1}{n}, & k\in P_i,\\[2pt]
p_{ik}, & k\notin P_i ,
\end{cases}
\label{eq:logit_grad}
\end{align}
where $P_i=\{k:Y_{ik}=1\}$, $|P_i|=n$. The parameter gradient is
$\nabla_\theta\ell_i=\sum_k(p_{ik}-\bar{y}_{ik})\,\nabla_\theta S_{ik}$. The
positive coordinates $k\in P_i$ (target $\frac{1}{n}$) pull together the $n$
cross-clothing matches of $i$'s identity---the invariance contribution---while
the negative coordinates $k\notin P_i$ push apart cross-identity look-alikes---the
discrimination contribution. Summing over queries and adding the symmetric
gallery-to-query term gives the additive split $h=h_{\mathrm{inv}}+h_{\mathrm{disc}}$, with $h_{\mathrm{inv}}$ supported on within-block
(positive) pairs and $h_{\mathrm{disc}}$ on off-block (negative) pairs.

\subsection{Convexity of the Allocation Objective}


We consider the continuous allocation problem
\begin{equation}
\min_{m,n>0}\;
\mathcal{E}(m,n)
=
\frac{\operatorname{tr}\Sigma_{\mathrm{B}}}{m}
+
\frac{\operatorname{tr}\Sigma_{\mathrm{W}}}{2n}
~
\text{s.t.}~
2mn\leq B_{\max}.
\end{equation}
Since the covariance matrices $\Sigma_{\mathrm{B}},\Sigma_{\mathrm{W}}\succeq \textbf{0}$,
their traces are nonnegative. We consider this non-degenerate
case where both traces are strictly positive. As $\mathcal{E}$ decreases when either $m$ or $n$ increases,
any minimizer must saturate the budget:
$2mn=B_{\max}$.
Thus, $m=B_{\max}/(2n)$, and the objective along the budget
frontier becomes
\begin{equation}
\widetilde{\mathcal{E}}(n)
=
\frac{2n\,\operatorname{tr}\Sigma_{\mathrm{B}}}{B_{\max}}
+
\frac{\operatorname{tr}\Sigma_{\mathrm{W}}}{2n},
\qquad n>0.
\end{equation}
Its derivatives are
\begin{equation}
\widetilde{\mathcal{E}}'(n)
=
\frac{2\operatorname{tr}\Sigma_{\mathrm{B}}}{B_{\max}}
-
\frac{\operatorname{tr}\Sigma_{\mathrm{W}}}{2n^2},
\qquad
\widetilde{\mathcal{E}}''(n)
=
\frac{\operatorname{tr}\Sigma_{\mathrm{W}}}{n^3}>0.
\end{equation}
Hence, $\widetilde{\mathcal{E}}$ is strictly convex on
$(0,\infty)$, so its unique stationary point is its unique
global minimizer. Solving
$\widetilde{\mathcal{E}}'(n^\star)=0$ gives
\begin{equation}
n^\star
=
\sqrt{
\frac{B_{\max}\operatorname{tr}\Sigma_{\mathrm{W}}}
     {4\operatorname{tr}\Sigma_{\mathrm{B}}}
},
~
m^\star
=
\sqrt{
\frac{B_{\max}\operatorname{tr}\Sigma_{\mathrm{B}}}
     {\operatorname{tr}\Sigma_{\mathrm{W}}}
}.
\end{equation}
Therefore,
\begin{equation}
\rho=
\frac{m^\star}{n^\star}
=
\frac{2\operatorname{tr}\Sigma_{\mathrm{B}}}
     {\operatorname{tr}\Sigma_{\mathrm{W}}},
\end{equation}
which yields Eq.~15 of the paper. The controller subsequently projects this continuous solution onto the feasible integer
allocations.

\subsection{Loss Function Analysis}
\label{sec:loss_function_analysis}
This section analyzes the contrastive loss formulation used in MINER to clarify its intended optimization behavior.
The loss is designed to explicitly control how gradient mass is distributed across multiple same-identity positives and identity-confusable negatives within a batch, a property that is critical for learning clothing-invariant representations.
We therefore examine the loss at the gradient level, relate it to prior supervised contrastive formulations, and show how its structure aligns with the discriminative pressure induced by our sampling strategy. Throughout this section we use the single-modal batch shape $\langle m,n\rangle$ of the main paper, where each identity contributes $n$ query and $n$ gallery images, so $N=mn$; the asymmetric cross-modal generalization is treated in \cref{sec:tvrid}.

\parhead{Gradient interpretation.}
\Cref{app:grad} derived the per-query gradient
$\partial \ell_i/\partial S_{ik}=\operatorname{softmax}(S_{ik})-\bar{y}_{ik}$
for the query-to-gallery direction $\mathcal{L}_{q\rightarrow g}$, where each query carries
exactly $n$ positive gallery samples in the block-diagonal supervision matrix $Y$, i.e.,
$\sum_{j} Y_{ij}=n$ for all $i$ and $\bar{y}_{ij}=Y_{ij}/n$. We read off its two regimes.
For a \textbf{positive} pair ($Y_{ik}=1$) the gradient is
\[
\frac{\partial \ell_i}{\partial S_{ik}}=\operatorname{softmax}(S_{ik})-\frac{1}{n},
\]
which drives the predicted softmax probability toward $1/n$, spreading probability mass
evenly across the $n$ cross-clothing matches of the identity. For a \textbf{negative} pair
($Y_{ik}=0$) it reduces to
\[
\frac{\partial \ell_i}{\partial S_{ik}}=\operatorname{softmax}(S_{ik}),
\]
so the update magnitude equals the predicted probability and is largest for the
highest-scoring negatives.

Because our batch construction explicitly concentrates identity-level hard negatives
through the soft sampling distribution of $\pi(p \mid p_s)$ of the main paper,
these negatives systematically attain larger similarity scores $S_{ik}$ early in training,
produce disproportionately large gradients, and dominate the optimization dynamics.
This provides a formal explanation for the strong discriminative pressure induced by our
mining strategy and its empirical impact on improving global retrieval metrics such as mAP.
At the same time, the block-diagonal positive structure distributes gradient mass across the
multiple cross-clothing positives of each identity, preventing collapse to single-instance
memorization and explicitly enforcing clothing-invariant identity clusters.

\parhead{Relation to Supervised Contrastive Learning.}
\citet{khosla2020supervised} identify two ways to generalize contrastive learning to arbitrary numbers of positives. We adopt the ``sum-outside-log'' formulation, defined as:
\begin{equation}
\label{eq:supcon_out}
\mathcal{L}^{\text{sup}}_{\text{out}}
\!=\!\sum_{i\in I}\!-\frac{1}{|P(i)|}
\!\sum_{p\in P(i)}
\!\log
\frac{\exp(\frac{\mathbf{z}_i^\top \mathbf{z}_p}{\tau})}
{\sum_{a\in A(i)} \exp(\frac{\mathbf{z}_i^\top \mathbf{z}_a}{\tau})} ,
\end{equation}
where $A(i)$ is the index set of candidates contrasted against anchor $i$, and $P(i)=\{p\in A(i)\mid \tilde y_p=\tilde y_i\}$ denotes the set of positives sharing the ground-truth identity label $\tilde{y}$ with the anchor.

\emph{Query$\rightarrow$Gallery.}
Consider a bipartite batch consisting of query embeddings $\{\mathbf{q}_i\}_{i=1}^{N}$ and
gallery embeddings $\{\mathbf{g}_j\}_{j=1}^{N}$, with $N=mn$.
In the query$\rightarrow$gallery direction, each query $\mathbf{q}_i$ serves as an anchor and is
compared against all gallery embeddings, which form the contrastive set.
Adopting the notation of Eq.~\ref{eq:supcon_out}, we set $I=\{1,\ldots,N\}$ and
$A(i)=\{1,\ldots,N\}$ with $(\mathbf{z}_i,\mathbf{z}_a)=(\mathbf{q}_i,\mathbf{g}_a)$.
The positive set for anchor $i$ is
$P(i)=\{\,j\in\{1,\ldots,N\}\mid Y_{ij}=1\,\}$; by construction $|P(i)|=n$ for all $i$.
Substituting into Eq.~\ref{eq:supcon_out} and averaging over the $N$ anchors yields
\begin{equation}
\label{eq:supcon_out_q2g}
-\frac{1}{N n}\sum_{i=1}^{N}\sum_{j:\,Y_{ij}=1}
\log\frac{\exp(S_{ij})}{\sum_{k=1}^{N}\exp(S_{ik})},
\end{equation}
which is identical to the query-to-gallery objective $\mathcal{L}_{q\rightarrow g}$
(Eq.~10 of the main paper).

\emph{Gallery$\rightarrow$Query.}
The gallery-to-query term is defined symmetrically by swapping the roles of queries and galleries, yielding the gallery-to-query objective $\mathcal{L}_{g\rightarrow q}$ (Eq.~11 of the main paper); the final objective averages both directions as in $\mathcal{L}_{\text{total}}$ (main paper, Sec.~3.2).

\parhead{Motivation for bipartite normalization}
\label{sec:loss_bipartite_motivation}

MINER trains on mined batches split into a query set $Q$ and a gallery set $G$, with
$|Q|=|G|=N=mn$ and each identity contributing $n$ images to each side. The bipartite
objective $\mathcal{L}_{q\rightarrow g}$/$\mathcal{L}_{g\rightarrow q}$ of the main paper) normalizes each query over galleries and each gallery
over queries. To understand the effect of this choice, we compare it to an all-to-all supervised contrastive variant that also
includes within-set candidates (query--query and gallery--gallery).

Fix a query anchor index $i$ with embedding $\mathbf{q}_i=f(q_i)$ and define the all-to-all candidate set
\[
\mathcal{A}^{\mathrm{all}}(i)\triangleq \{\mathbf{g}_j\}_{j=1}^{N}\cup(\{\mathbf{q}_k\}_{k=1}^{N}\setminus\{\mathbf{q}_i\}),
\]
where $\mathbf{g}_j=f(g_j)$ and $\mathbf{q}_k=f(q_k)$ are embeddings of the gallery and query images
$\mathcal{G}=\{g_j\}_{j=1}^{N}$ and $\mathcal{Q}=\{q_k\}_{k=1}^{N}$, respectively.
Define the all-to-all probability over candidates $\mathbf{u}\in\mathcal{A}^{\mathrm{all}}(i)$ by
\begin{equation}
p^{\mathrm{all}}_{i\to \mathbf{u}}
\triangleq
\frac{\exp(\mathbf{q}_i^\top \mathbf{u}/\tau)}
{\sum_{\mathbf{v}\in \mathcal{A}^{\mathrm{all}}(i)} \exp(\mathbf{q}_i^\top \mathbf{v}/\tau)}.
\end{equation}
Two effects distinguish bipartite from all-to-all normalization; we lead with the one
that moves the objective's optimum rather than merely its scale.

\emph{Target-mass dilution (primary).}
Under bipartite normalization each query distributes unit positive-target mass over its
$n$ cross-set (gallery) positives, so each receives $1/n$ (\cref{app:grad}). Under the
all-to-all objective the anchor's positive set additionally includes its $(n-1)$
same-identity queries, for a total of $N_p=2n-1$ positives, so each cross-set positive
now receives only $1/N_p$. The total target mass placed on the retrieval-relevant
cross-set positives therefore drops from $1$ to
\begin{equation}
\label{eq:target_mass}
\frac{n}{2n-1}\;\xrightarrow[\;n\to\infty\;]{}\;\frac{1}{2}.
\end{equation}
At MINER's operating points this is already near the limit: $17/33\approx0.52$ at the
PRCC optimum $\langle 5,17\rangle$ and $4/7\approx0.57$ at the Celeb-L optimum
$\langle 21,4\rangle$. Roughly half of the supervisory target mass is thus diverted to
within-set query--query pairs, away from the query--gallery comparisons that retrieval
actually evaluates. Unlike a uniform rescaling, this reweighting moves the fixed point of
the loss, not merely its magnitude.

\emph{Probability-mass shrinkage (secondary).}
The same within-set candidates also rescale the model's predicted distribution. Partition
the all-to-all denominator into cross-set and within-set terms:
\begin{equation}
\begin{aligned}
D_i^{q\to g} &\triangleq \sum_{j=1}^{N}\exp(\mathbf{q}_i^\top \mathbf{g}_j/\tau),\\
D_i^{q\to q} &\triangleq \sum_{k\neq i}\exp(\mathbf{q}_i^\top \mathbf{q}_k/\tau).
\end{aligned}
\end{equation}
Then, for any gallery embedding $\mathbf{g}_j$,
\begin{equation}
\label{eq:alpha_scale}
\begin{aligned}
p^{\mathrm{all}}_{i\to \mathbf{g}_j}
&=
\frac{\exp(\mathbf{q}_i^\top \mathbf{g}_j/\tau)}{D_i^{q\to g}+D_i^{q\to q}}
= \alpha_i\,p^{q\to g}_{ij},\\
p^{q\to g}_{ij}
&\triangleq
\frac{\exp(\mathbf{q}_i^\top \mathbf{g}_j/\tau)}{D_i^{q\to g}},\\
\alpha_i
&\triangleq \frac{D_i^{q\to g}}{D_i^{q\to g}+D_i^{q\to q}} \;<\; 1.
\end{aligned}
\end{equation}
Adding query--query candidates shrinks \emph{all} cross-set probabilities by the common
factor $\alpha_i$ whenever $D_i^{q\to q}$ is non-negligible, so the total probability mass
allocated to \emph{all} gallery candidates becomes
\begin{equation}
\label{eq:gallery_mass_alpha}
\sum_{j=1}^{N} p^{\mathrm{all}}_{i\to \mathbf{g}_j}
=
\alpha_i,
\end{equation}
whereas under bipartite normalization $\sum_{j=1}^{N} p^{q\to g}_{ij}=1$. Taken alone, this
is a uniform attenuation of the cross-set gradient---equivalent to a smaller effective step
size---and is therefore milder than the target-mass effect above; we include it because it
compounds with that effect under the same within-set candidates.

\emph{Why the within-set positives are inert.}
One might expect the $(n-1)$ within-set positives to compensate by contributing useful
gradient. They do not. Under $\ell_2$-normalization and strong initialization, same-identity
query pairs become nearly collinear: for such a pair the tangential component has norm
\begin{equation}
\label{eq:tangent_small}
\begin{aligned}
\bigl\lVert \mathbf{q}_k-(\mathbf{q}_i^\top \mathbf{q}_k)\mathbf{q}_i \bigr\rVert_2
&= \sqrt{1-(\mathbf{q}_i^\top \mathbf{q}_k)^2},\\[-2pt]
&\approx 0 \quad \text{when } \mathbf{q}_i^\top \mathbf{q}_k \approx 1 .
\end{aligned}
\end{equation}
To make this precise, we decompose the all-to-all positive gradient into cross-set and
within-set contributions. Let $P^{\mathrm{all}}(i)\subseteq\mathcal{A}^{\mathrm{all}}(i)$ denote
the positive set of $\mathbf{q}_i$ and let $N_p\triangleq|P^{\mathrm{all}}(i)|$. Define the
per-anchor all-to-all loss
\[
\ell_i^{\mathrm{all}}
\triangleq
-\frac{1}{N_p}\sum_{\mathbf{u}\in P^{\mathrm{all}}(i)} \log p^{\mathrm{all}}_{i\to \mathbf{u}} .
\]
Under the MINER sampler (Alg.~1 of the main paper),
$P^{\mathrm{all}}(i)$ contains $n$ cross-set positives and $(n-1)$ within-set positives, hence
$N_p=2n-1$. For compactness, define the positive index subsets
\[
P^{\mathrm{all}}_{g}(i)\triangleq P^{\mathrm{all}}(i)\cap \mathcal{G},
\qquad
P^{\mathrm{all}}_{q}(i)\triangleq P^{\mathrm{all}}(i)\cap \mathcal{Q}.
\]
Write $\mathbf{q}_i=\tilde{\mathbf{q}}_i/\lVert\tilde{\mathbf{q}}_i\rVert_2$ and define
\[
\begin{aligned}
c_i &\triangleq \frac{1}{\tau\lVert\tilde{\mathbf{q}}_i\rVert_2},\\
\Delta^{q\to g}_{ij} &\triangleq \mathbf{g}_j-(\mathbf{q}_i^\top \mathbf{g}_j)\mathbf{q}_i,\\
\Delta^{q\to q}_{ik} &\triangleq \mathbf{q}_k-(\mathbf{q}_i^\top \mathbf{q}_k)\mathbf{q}_i.
\end{aligned}
\]
Then the positive contribution decomposes as
\begin{equation}
\label{eq:pos_grad_query_style}
\begin{aligned}
\left.\frac{\partial \ell^{\mathrm{all}}_i}{\partial \tilde{\mathbf{q}}_i}\right|_{\text{pos}}
=&
c_i\Bigg(
\sum_{\mathbf{g}_j\in P^{\mathrm{all}}_{g}(i)}
\Delta^{q\to g}_{ij}\Bigl(p^{\mathrm{all}}_{i\to \mathbf{g}_j}-\tfrac{1}{N_p}\Bigr) \\
&+\sum_{\mathbf{q}_k\in P^{\mathrm{all}}_{q}(i)}
\Delta^{q\to q}_{ik}\Bigl(p^{\mathrm{all}}_{i\to \mathbf{q}_k}-\tfrac{1}{N_p}\Bigr)
\Bigg).
\end{aligned}
\end{equation}
When $\mathbf{q}_i^\top \mathbf{q}_k \approx 1$ (Eq.~\ref{eq:tangent_small}),
$\lVert\Delta^{q\to q}_{ik}\rVert_2\approx 0$, so the second sum is negligible and the update
direction is dominated by cross-set terms---while these within-set logits still inflate
$D_i^{q\to q}$ and, through $\alpha_i$ (Eq.~\ref{eq:alpha_scale}), drain the magnitude budget
available to cross-set terms. The within-set positives thus pay the full dilution cost of
both effects above and return almost no gradient in exchange.

\emph{What about within-set negatives?}
A natural objection is that query--query \emph{negatives} (different-identity query pairs)
supply discriminative signal that bipartite normalization discards. In this setting they are
redundant: because every identity contributes images to \emph{both} $Q$ and $G$, the contrast
between any two identities is already exercised by cross-set query--gallery negatives in $S$.
Each off-block entry $S_{ij}$ with $\mathrm{pid}(q_i)\neq\mathrm{pid}(g_j)$ is exactly such a
hard-negative comparison, and the MINER sampler selects identities so that these off-block
pairs are maximally confusable. Within-set negatives would re-encode the same
identity-separation pressure on a different pairing whose utility is dataset- and
regime-dependent, at the cost of the dilution quantified above.

Together, these observations explain why bipartite normalization matches MINER's retrieval
setting: it preserves probability mass and uniform-positive target mass on the cross-set
comparisons selected by the MINER sampler (Alg.~1 of the main
paper) and evaluated at test time, while discarding within-set pairs that are either inert
(positives) or redundant (negatives).